%% file: main.tex
\documentclass[runningheads]{llncs}
\usepackage{graphicx}
\usepackage{tikz}
\usepackage{comment}
\usepackage{amsmath,amssymb} % define this before the line numbering.
\usepackage{color}
\usepackage[utf8]{inputenc} % allow utf-8 input
\usepackage[T1]{fontenc}    % use 8-bit T1 fonts
\usepackage{url}            % simple URL typesetting
\usepackage{booktabs}       % professional-quality tables
\usepackage{amsfonts}       % blackboard math symbols
\usepackage{nicefrac}       % compact symbols for 1/2, etc.
\usepackage{microtype}      % microtypography
\usepackage{xcolor}         % colors
\usepackage{hhline}%#######
\usepackage{bbm}
\usepackage{boldline}
\usepackage{graphicx}
\usepackage{wrapfig}
\usepackage{multirow}
\usepackage[font=small]{caption}
\usepackage{booktabs}
\usepackage{subcaption}
\usepackage{caption}
\usepackage{diagbox}
\usepackage{hyperref}
\hypersetup{
    colorlinks=true,
    urlcolor=blue,
}
\usepackage{cite}
\usepackage{calc}
\newlength{\arrayrulewidthOriginal}

\usepackage[accsupp]{axessibility}  % Improves PDF readability for those with disabilities.

\begin{document}
% \renewcommand\thelinenumber{\color[rgb]{0.2,0.5,0.8}\normalfont\sffamily\scriptsize\arabic{linenumber}\color[rgb]{0,0,0}}
% \renewcommand\makeLineNumber {\hss\thelinenumber\ \hspace{6mm} \rlap{\hskip\textwidth\ \hspace{6.5mm}\thelinenumber}}
% \linenumbers
\pagestyle{headings}
\mainmatter
\def\ECCVSubNumber{6282}  % Insert your submission number here
\newcommand{\JP}[1]
{
	\textcolor{blue}{\bfseries{JP: {#1}}}
}
\newcommand{\WJ}[1]
{
	\textcolor{red}{\bfseries{WJ: {#1}}}
}
\newcommand{\JH}[1]
{
	\textcolor{cyan}{\bfseries{JH: {#1}}}
}
\newcommand{\Skip}[1]
{
}
\title{Tailoring Self-Supervision for \\Supervised Learning} % Replace with your title

% INITIAL SUBMISSION 
\begin{comment}
\titlerunning{ECCV-22 submission ID \ECCVSubNumber} 
\authorrunning{ECCV-22 submission ID \ECCVSubNumber} 
% \author{Anonymous ECCV submission}
% \institute{Paper ID \ECCVSubNumber}
\author{
WonJun Moon, Ji-Hwan Kim, Jae-Pil Heo\thanks{Corresponding author}}
\institute{Sungkyunkwan University}
\end{comment}
%******************

% CAMERA READY SUBMISSION
% \begin{comment}
\titlerunning{Tailoring Self-Supervision for Supervised Learning}
% If the paper title is too long for the running head, you can set
% an abbreviated paper title here
%
% \author{WonJun Moon\inst{1}\orcidID{0000-0003-2805-0926} \and
% Ji-Hwan Kim\inst{2,3}\orcidID{1111-2222-3333-4444} \and
% Jae-Pil Heo\inst{3}\orcidID{2222--3333-4444-5555}}
\author{WonJun Moon \and
Ji-Hwan Kim \and
Jae-Pil Heo\thanks{Corresponding author}\\}
\authorrunning{WJ. Moon et al.}
% First names are abbreviated in the running head.
% If there are more than two authors, 'et al.' is used.
%
\institute{Sungkyunkwan University\\
\email{\{wjun0830,damien,jaepilheo\}@skku.edu}}
% \institute{Sungkyunkwan University, Princeton NJ 08544, USA \and
% Springer Heidelberg, Tiergartenstr. 17, 69121 Heidelberg, Germany
% \email{lncs@springer.com}\\
% \url{http://www.springer.com/gp/computer-science/lncs} \and
% ABC Institute, Rupert-Karls-University Heidelberg, Heidelberg, Germany\\
% \email{\{abc,lncs\}@uni-heidelberg.de}}
% \end{comment}
%******************
\maketitle
\begin{abstract}
\input{_0_abstract}
\end{abstract}
%%%%%%%%% BODY TEXT
\input{_1_introduction}
\input{_2_related_work}
% \input{3-1_properties}
\input{_3_method}
\input{_4_experiments}

\input{_5_further_analysis}
\input{_6_conclusion}
\bibliographystyle{splncs04}
\bibliography{egbib}
\newpage
\input{Supplementary/abstract}
\input{Supp_lambdaExp}
\input{Supp_patchsizeExp}

\input{Supp_GAP}
\input{Supp_SupCLRFullOOD}
\input{Supp_ImbalancedClassification}
\input{Supp_OODanalysis}
\input{Supp_implementationDetails}

\end{document}

%% file: _0_abstract.tex
Recently, it is shown that deploying a proper self-supervision is a prospective way to enhance the performance of supervised learning. 
Yet, the benefits of self-supervision are not fully exploited as previous pretext tasks are specialized for unsupervised representation learning.
To this end, we begin by presenting three desirable properties for such auxiliary tasks to assist the supervised objective.
First, the tasks need to guide the model to learn rich features. Second, the transformations involved in the self-supervision should not significantly alter the training distribution. 
Third, the tasks are preferred to be light and generic for high applicability to prior arts. 
Subsequently, to show how existing pretext tasks can fulfill these and be tailored for supervised learning, we propose a simple auxiliary self-supervision task, predicting localizable rotation (\textbf{LoRot}).
% Unlike the popular rotation prediction task, we rotate a patch of an image and ask the model to predict the rotation degree. 
% This encourages the model to learn rich features by dealing with various object parts rotated.
% Furthermore, we show that rotating a patch does not incur significant data distribution shift which enables an effective and efficient learning process by adopting a multi-task learning strategy.
% As a result, we drastically decrease the cost of applying pretext tasks in supervised learning.
Our exhaustive experiments validate the merits of LoRot as a pretext task tailored for supervised learning in terms of robustness and generalization capability.
Our code is available at \href{https://github.com/wjun0830/Localizable-Rotation}{https://github.com/wjun0830/Localizable-Rotation}.
\keywords{Pretext task, Auxiliary self-supervision, Supervised learning}

%% file: _1_introduction.tex
\section{Introduction}
\label{section:Grid Rotation}
% Beyond the success in visual recognition without human supervision\cite{yang2016joint, caron2018deep, wu2018unsupervised, chen2020learning, feng2019self}, there has been attempts to adopt self-supervision to supervised learning.
Beyond the success in visual recognition without human supervision~\cite{yang2016joint, caron2018deep, wu2018unsupervised, chen2020learning, feng2019self, zhai2019large}, there have been attempts to adopt self-supervision to supervised learning.
Pioneering methods demonstrated that self-supervision indeed improves robustness along with human guidance~\cite{carlucci2019domain, hendrycks2019using, perera2020generative}.
They utilized self-supervision as an auxiliary task to support the feature learning.
However, as these methods utilize the existing self-supervisions specialized for unsupervised representation learning, the benefits of self-supervision are restrained.
Specifically, self-supervision is generally designed for representation learning which is performed better by the primary objective of supervised learning.
Furthermore, employed transformations for pretext tasks often trigger significant data distribution shifts.
% \WJ{Specifically, representation learning, the commonly granted role for self-supervision, is performed with supervision, and employed transformations for pretext tasks often trigger significant data distribution shift.}
For instance, rotation task, the most popular self-supervision in supervised domain, is not an exception since learning rotation-invariant features barely help the primary task~\cite{lee2020self, gidaris2018unsupervised}.
In fact, rotation task only achieves insignificant gains or sometimes even degrades the performance when applied in the form of multi-task learning or as an augmentation technique~\cite{lee2020self, chen2020simple}.
% In fact, rotation task even degrades the performance when applied by the form of multi-task learning or as an augmentation technique~\cite{lee2020self, chen2020simple}.
This motivates us to develop a complementary pretext task to supervised objectives.
% However, these methods only exploit the existing self-supervised techniques that are designed for unsupervised representation learning.
% Indeed, representation learning is usually performed with labeled datasets in supervised learning.
% To be specific, conventional self-supervised methods are designed for general feature learning which is usually performed with labeled datasets in supervised learning.
% As such, although they highlight a great potential of self-supervision, designing a schema for pretext tasks for supervised learning has not been studied seriously.

% in spite of their importance.
% For instance, although the rotation task~\cite{gidaris2018unsupervised} was renowned for its success in unsupervised learning, it requires much cost to achieve limited performance gain in supervised learning.
% Previously, rotation~\cite{gidaris2018unsupervised} was renowned for its success in unsupervised learning.
% However, due to its limitations in three aspects discussed, 

Therefore, in this paper, we first introduce three desirable properties of auxiliary self-supervision to maximize the benefits in supervised learning: 1) learning rich representations, 2) maintaining data distribution, 3) providing high applicability. First, the pretext task should guide the model to learn complementary features with the original ones from supervised learning. Models trained only with a primary task such as the classification often focuses on most discriminative parts of objects where such features provide shortcuts to solve the problem~\cite{nguyen2015deep, geirhos2020shortcut}. Thus, the primary goal of an auxiliary task is to help the model to capture additional detailed features, which are known to improve the robustness of the model such as detecting out-of-distribution samples as well as its accuracy~\cite{hendrycks2019using}.
Second, the transformation itself should not bring significant data distribution shifts.
Although an ideal convolutional neural network (CNN) should be invariant to the transformations such as translation or rotation\cite{mallat2016understanding}, in realistic circumstances, the shift of global views of images is often known to be harmful to classification tasks\cite{tack2020csi, chen2020simple}.
% Therefore, distribution shift should be considered when designing transformation function for ideal CNN.
% Specifically, the shift of global views of images are known to be harmful to classification tasks while suitable for generating negative instances representing unknown samples~\cite{tack2020csi, chen2020simple}.
% Specifically, the shift of global views of images are known to be harmful to classification tasks while suitable for representing out-of-distribution samples~\cite{tack2020csi, chen2020simple}.
% not suitable since it is known to hinders the primary objective (e.g. Image Classification), so thus it can degrade the generalization performance~\cite{tack2020csi, chen2020simple}.
% 당연한 명제이기보단 observe 한다.
% Moreover, the transformation of a proper amount of perturbation could provide a data augmentation effect.
Third, the pretext task is preferred to be highly applicable to existing model architectures and strategies of supervised learning in terms of the computational overhead and the amount of modification.

To validate the importance of these properties, we propose an auxiliary self-supervision task tailored for supervised learning, Localizable Rotation (LoRot), which forms the localization quizzes by rotating only a part of an image.
Note that we choose rotation task to be modified into a tailored version for supervised domain on behalf of other pretext tasks due to its effectiveness of localizing the salient objects following previous works~\cite{hendrycks2019using, lee2020self}.
% In this regard, we propose a tailored auxiliary task for supervised learning, Localizable Rotation (LoRot), which forms the localization quizzes by rotating only the part of an image. 
LoRot provides complementary benefits to supervised learning since the model should first localize the patch to solve the rotation task.
Specifically, it encourages the model to learn rich features for rotational clues within a part of the image even if they are less discriminative for the supervised objectives.
% as the location of the patch randomly moves, it is forced to find tangible objects in every spatial location. 
% it is encouraged to learn rich features for rotational clues even if they are less discriminative for the supervised objectives
% In this regard, we propose a tailored auxiliary task for supervised learning, Localizable Rotation (LoRot), in which the neural network should first localize the patch to solve the rotation task.
% LoRot benefits for supervised learning by rotating only a part of an image; as the location of the patch randomly moves, it is forced to find tangible objects in every spatial locations. 
Furthermore, we found that rotating a small patch does not incur a significant data distribution shifts. % which allows LoRot to retain the advantage of data augmentation.
% 너무 당연한것처럼 얘기함. 실험적인건데.
Finally, the LoRot requires small extra computational costs and implementation efforts, since it is designed for multi-task learning that only requires one additional classifier. 
In our extensive experiments, we validate that LoRot is effective at boosting the robustness and generalization capability of supervised models and even provides state-of-the-art results.
Specifically, we evaluate LoRot on various tasks including out-of-distribution (OOD) detection, imbalanced classification, adversarial attack, image classification, localization, and transfer learning in Sec.~\ref{section:Experiments}.

\Skip{
            
            % There has been a drastic progress in computer vision including object detection, semantic segmentation and image classification. Behind the fact that their performances are now comparable to the one of actual human, they require expensive supervision from human. To overcome this limitation, there has been much effort on unsupervised representation learning also known as self-supervised learning. Self-supervision can be categorized into two classes depending how pretext tasks are designed: relation-based and transform-based which relation-based utilize relation between augmented samples and transform-based predicts transformation of the image.
            %(computer비전의 발전 -> unsupervised learning전개)
            
            %
            % Introduction of the field with computer vision and unsupervised learning
            %
            There has been a drastic progress in computer vision including image classification, semantic segmentation, and object detection. 
            Behind the fact that their performances are now comparable to the one of actual humans, they require expensive supervision from humans. 
            To overcome this limitation, there has been sustained effort into unsupervised representation learning.
            % Unsupervised learning basically seeks for hidden patterns from data without any human supervision.
            As one promising stream of unsupervised learning, self-supervision has been getting a great deal of attention in recent years.
            Self-supervision produces supervision from data as itself to effectively learn their visual patterns.
            Remarkable advances in self-supervised learning have realized learning discriminative representations for images and videos without any human labor.
            
            %
            % Self-supervised learning
            %
            % Recently, most self-supervised approaches fall into two classes; designing pretext tasks from multiple images or from a single image.
            Recently, most self-supervised approaches fall into two classes; relation-based and transform-based methods.
            Relation-based models~\cite{chen2020simple, grill2020bootstrap, he2020momentum} maximize the agreement among positive instances, and minimize it from negatives.
            They define positive samples as different views from an identical image, and negative ones as those from other samples.
            On the other hand, transform-based methods~\cite{dosovitskiy2014discriminative, noroozi2016unsupervised,gidaris2018unsupervised} deploy transformations like rotation, and learn to predict the details of transformation such as the degree of rotation.
            As transformation is conducted within one single image, they do not require a large set of negative samples, resulting in high training efficiency.
            % For rest of the paper, we simply call these methods transformation-based in this paper.
            
            %
            % Self-supervision for supervised learning
            %
            In spite of its popularity in unsupervised learning, self-supervision has not been explored much for supervised learning.
            When the pretext task of self-supervision is given along with the primary task of supervised learning, it can provide complementary benefits like the improved robustness to unknown samples.
            % Especially for single-image self-supervision methods, their performance gain is limited due to the properties of transformations they use.
            % Especially for single-image self-supervision methods, however, the performance gain is still limited due to the properties of transformations they use.
            % However, along with multi-image based self-supervision's modification~\cite{khosla2020supervised, tack2020csi},
            % pioneering methods~\cite{hendrycks2019using, lee2020self} have shown that single-image based self-supervision, particularly rotation, can help to improve the performance of learning with labels as well as the robustness.%가능성
            Pioneering methods~\cite{khosla2020supervised, tack2020csi, lee2020self, hendrycks2019using} have shown that self-supervision indeed improves the performance of both generalization and robustness along with human supervision.
            Although they spotlight a great potential of self-supervision, transformations they deploy have not been studied seriously for supervised learning in spite of their importance in self-supervision.
            % The limitations of the single-image based approaches are mainly from the data-distribution shift~\cite{tack2020csi} caused by the transformation they use.
            % That is, in supervised settings, the shift of global semantics of images is not necessary since the primary objective of supervised learning achieves that goal.
            % Therefore, transformations of the previous single-image based methods for unsupervised learning could even degrade the generalization performance in supervised learning.
            
            %
            % Requirements of self-supervision for supervised learning
            %
            % data augmentation 표현 바꾸기
            % In this paper, we point out the limitations of the current single image-based self-supervision in supervised learning by proposing two factors that should be considered for the adaptation; 1) data distribution, 2) richer representations.
            In this paper, we point out the limitations of the current self-supervision transformations by proposing three necessary conditions for supervised learning: 1) learning rich representations, 2) reducing data-distribution shift, 3) providing flexiblity.
            First, the transformation itself should not shift the data distribution but introduce hidden samples in current data-distribution. 
            To be specific, the shift of global views of images should be prohibited since it hinders the primary objective of supervised learning, degrading the generalization performance. 
            When transformation meets the above condition, naturally, it can give a performance gain since it plays a role of data augmentation.
            Second, the pretext task of self-supervision should give a complementary effect to the original objective of supervised learning.
            Models with a primary task like classification often achieve great performance through relatively short paths like focusing on only textures or small discriminative parts of an object.
            In this case, the complementary goal of pretext tasks could be enforcing the model to gather surplus details to perceive the whole object.
            These affluent representations should improve the robustness to out-of-distribution and imbalanced samples as well as boosting the generalization capability.
            Finally, the pretext task should not force a large modification or additional computational cost to be widely applied in supervision tasks.
            
            % In this paper, we point out the limitations of the current transformations by proposing two necessary characteristics of self-supervision for supervised learning; 1) data augmentation, 2) rich representations.
            % First, the transformation itself should play the role of data augmentation.
            % It means that transformation should not make an extremely different views from the original one, which can cause data-distribution shift.
            % The shift of global views of images is prohibited since it hinders the primary objective of supervised learning, which can even degrade the generalization performance.
            % When transformation meets the above condition, it can give a natural performance gain as itself since it can augment data.
            % Second, the pretext task of self-supervision should give a complementary effect to the original objective of supervised learning.
            % Models with a primary task like classification often achieve great performance through relatively short paths like focusing on only textures or small parts of an object.
            % In this case, the complementary goal of pretext tasks could be enforcing the model to learn the details of the other object parts.
            % These rich representations should improve the robustness to out-of-distribution and imbalanced samples as well as boosting the generalization capability.
            
            %
            % Our approach
            %
            % \WJ{To satisfy the necessary conditions, we propose a novel paradigm of supplementary self-supervision and come up with PatchRot.}
            To successfully the achieve these properties, we propose a novel self-supervision, PatchRot, which is a rotation-based method with random patches.
            Unlike the previous work of rotation~\cite{gidaris2018unsupervised}, PatchRot rotates only a part of an image rather than the whole to maintain the data in its distribution.
            When we rotate a patch and keep other parts intact, the data distribution is not shifted so that the primary objective can be achieved properly.
            Furthermore, we suggest an additional and effective constraint for forming patches, which makes square patches in a grid without overlap.
            This way spurs the model to focus on the orthogonal objective of self-supervision to enrich representation.
            Finally, all these processes can be done without requiring much extra cost for both computation and modification of base frameworks which makes PatchRot broadly applicable.
            % This way maximizes the orthogonal objective of self-supervision by encouraging the model to focus on more details of objects.
            In our extensive experiments of various tasks, we validate that our method shows the state-of-the-art results in terms of the robustness and generalization performance.
            Moreover, we show the complementary benefits of our method to data-augmentation and contrastive-learning approaches.
            % \JH{We need to add more explanation about the first try of supplementary manner.}
            % To the best of our knowledge, we are the first to explore and devise the pretext task in supplementary manner.

            % The reasons and introducing questions/problems
            
            % Inspired by these work, we study transform-based self-supervision under supervised setting and specifically, investigate two following questions:
            % What characteristic does single-image self-supervision need to be applied in unsupervised/supervised setting?
            % Can we improve the effectiveness of single-image self-supervision in supervised settings by considering the supervised learning’s objective?
            
            % Propose our method
            
            % Contributions
            
            % Based on the analysis of above questions, we design GridRot that is very simple and powerful.
            % Contribution 
            % 1. Considering the objectives of unsupervised and supervised learning, we study what characteristics should self-supervision have in each settings.
            % 2. We propose simple but effective self-supervision that retains not only the strength of self-supervision but also the one of augmentation: robustness and generalization.
            % 3. We evaluate the efficacy on Out-of-Distribution, Open-Set Recognition and Image Classification and achieve new state-of-the-art for many datasets.

}

%% file: _2_related_work.tex
\section{Related Work}
\textbf{Self-Supervised Learning.}
Self-supervised learning has received considerable attention in past years. Its typical objective is to learn general features through solving pretext tasks.
%To achieve this, pretext tasks are designed to be solved only when the neural network is capable of extracting general features from the image.
According to the number of instances to define pretext tasks, we can categorize the self-supervision into two groups, relation- and transformation-based ones. Relation-based approaches learn features to increase the similarity among a sample~\cite{chen2021exploring, he2020momentum, chen2020simple, caron2020unsupervised, tian2021understanding} and its transformed positive instances while some also treat other training samples as negative instances. The memory bank~\cite{he2020momentum, misra2020self} and in-batch~\cite{ye2019unsupervised, chen2020simple} samplings are notable negative instance selection techniques. 
% In contrast, there also have been approaches to reduce the size of the negative sample pool by utilizing siamese networks~\cite{caron2020unsupervised, grill2020bootstrap} or adding a relation module~\cite{patacchiola2020self}. 
In contrast, there have been approaches to only use the positive pairs with siamese networks~\cite{caron2020unsupervised, grill2020bootstrap} or adding a relation module~\cite{patacchiola2020self}. 
%Recently, relation-based self-supervisions are spotlighted as they are not only comparable but sometimes outperform fully supervised counterparts in some tasks.
Moreover, transform-based self-supervision is another main stream of representation learning that substantial efforts are made.
Remarkable methods are generating surrogate classes with data augmentation~\cite{dosovitskiy2014discriminative}, predicting the relative position of patches~\cite{doersch2015unsupervised}, solving jigsaw puzzle~\cite{noroozi2016unsupervised,kim2018learning,noroozi2018boosting, mundhenk2018improvements, chen2021jigsaw}, and predicting the degree of rotation~\cite{gidaris2018unsupervised}. LoRot also belongs to transform-based self-supervision but is devised for a different objective: to assist supervised learning.

Meanwhile, there have been new attempts to transfer the benefits of self-supervisions to supervised learning. SupCLR~\cite{khosla2020supervised} modified the relation-based self-supervision framework to directly take advantage of labeled data since the class labels clearly define both positive and negative instances.
Moreover, self-label augmentation (SLA) augmented the label space based on the Cartesian product of the supervised class label set and the data transformations label set because learning auxiliary pretext task degrades the performance~\cite{lee2020self}.
% According to SLA, adding the rotation task\cite{gidaris2018unsupervised} to the supervised learning as a multi-task objective is inefficient, since learning rotation-invariant features is hard but barely helps the primary task. This motivates us to develop a complementary self-supervised task for supervised models that can be easily attached to the existing framework as a multi-task objective. 
In contrast, LoRot is an adequate self-supervision for supervised learning that can be directly applied to existing methods. Further discussion is in Sec.~\ref{section:properties}.

%We share the motivation with latter approach that transform-based self supervision's potential is not being elicited in supervised setting. 
%However, self-label augmentation~\cite{lee2020self} solved this issue by altering the strategy of applying rotation. On the other hand, PatchRot is newly modified self supervision which is complementary with fully supervised method.

\textbf{Regional Data Transformation.}
Data augmentation is one of the most popular ways to improve classification accuracy~\cite{cubuk2019autoaugment, lim2019fast, zhang2017mixup}. 
%Among them, removing or replacing a certain region of an image is similar to our method in that we rotate a small portion of each image.
Among them, we introduce methods that modify local regions of an image. 
Cutout~\cite{devries2017improved} and random erasing~\cite{zhong2020random} randomly mask out square regions of input, while Cutmix~\cite{yun2019cutmix} cuts and pastes rectangular regions from other samples. 
LoRot shares the property of editing the local patch with them, however, the main difference is that our transformation retains all information within the image and is a solvable task.
% \WJ{LoRot shares the property of editing the local patch with them, however, the main difference is that our transformation is to create novel pretext task to improve not only the accuracy but also the robustness of supervised models.}
% main difference 새로운거 찾아야함. 목적, implementation 등 차이가 확연하게.
% LoRot shares the property of editing the local information with them, however, the main difference is that LoRot is an auxiliary task to improve not only the accuracy but also the robustness of supervised models.

%% file: _3_method.tex
\section{Methodology}
% In this section, we first discuss three desired properties for auxiliary self-supervision. Then, considering these details, we craft LoRot(Localizable Rotation) in two ways: explicit and implicit localization tasks, both tailored for supervised learning. In ~\cref{section:properties}, we discuss three preferred properties of auxiliary self-supervision for supervised learning. Then, in ~\cref{subsection:methodology} we introduce LoRot in two forms: LoRot with explicit and implicit localization tasks.
% We first discuss three desired properties for the auxiliary self-supervision for supervised learning in Sec.~\ref{section:properties}. Based on those, we introduce and discuss LoRot (Localizable Rotation) in two ways; explicit and implicit localization tasks, both tailored for supervised learning in Sec.~\ref{subsection:methodology}.
We first discuss three desired properties for the auxiliary self-supervision for supervised learning in Sec.~\ref{section:properties}. Based on those, we introduce and discuss two forms of LoRot (Localizable Rotation): having explicit and implicit localization tasks, both tailored for supervised learning in Sec.~\ref{subsection:methodology}.
\subsection{Desired Properties in Supervised Learning}
\label{section:properties}

In this section, we discuss three preferred properties of auxiliary self-supervision for supervised learning: (i) extracting rich representation, (ii) maintaining %%JP restraining
distribution, and (iii) high applicability, and point out the limitations of previous self-supervised methods in the manner of supervised learning.

In typical training of CNN for the classification task, the model tends to focus on identifying class-specific highly discriminative features to reach a high training accuracy. However, these features usually cover a limited portion of the objects since other parts can be unnecessary to achieve high training accuracy. This phenomenon is often called shortcut learning~\cite{geirhos2020shortcut}. It can be problematic when the model faces samples that do not belong to known classes but have the learned discriminative features~\cite{nguyen2015deep}. In such cases, discovering rich features including detailed parts of objects can enhance the robustness of the model~\cite{perera2020generative}. Auxiliary tasks for supervised learning can encourage the model to learn such less discriminative features with the complementary objectives~\cite{baxter1997bayesian, ruder2017overview}. For instance, the popular rotation prediction task~\cite{gidaris2018unsupervised} can spread the attention of the model toward object parts for predicting the degree of rotation. 
However, discriminative clues for predicting rotation degrees also exist, e.g., location and orientation. 
Indeed, the rotation prediction task requires the model to focus on high-level object parts, which are roughly the same image regions as the supervised classification task~\cite{gidaris2018unsupervised}.
So thus, the supervised learning with the auxiliary rotation task is still limited to identifying the most discriminative parts for both tasks.
% \cline{2-2}
%                               & CIFAR10 &  &  &  \\ \cline{1-2}
% \multicolumn{1}{|l|}{Baseline} & 95.01   &  &  &  \\
% \multicolumn{1}{|l|}{+Rot(DA)} & 92.76   &  &  &  \\
% \multicolumn{1}{|l|}{+Rot(MT)} & 93.38   &  &  &  \\ \cline{1-2}
\begin{table}[!t]
    \centering
    \small
    \begin{minipage}[t!]{0.42\linewidth}%\centering
    \setlength{\tabcolsep}{3pt} % Default value: 6pt
    \renewcommand{\arraystretch}{0.7} % Default value: 1
        	\centering
        	{\small
        	\caption{CIFAR10 classification accuracy (\%) with rotation using different strategies.}
        		\label{table_DAMTSS}
        		\begin{tabular}{l  c}
        		    \hlineB{2.5}
        		     & Accuracy \\ 
        		    \hline
        		    \multicolumn{1}{l|}{Baseline} & 95.01 \\ 
        		    \multicolumn{1}{l|}{+Rot (DA)} & 92.76 \\ 
        		    \multicolumn{1}{l|}{+Rot (MT)} & 93.38 \\
                    \hlineB{2.5}
        		\end{tabular}
        	}
    \end{minipage}\hfill%
    \begin{minipage}[t!]{0.54\linewidth}
    \setlength{\tabcolsep}{4pt} % Default value: 6pt
    \renewcommand{\arraystretch}{0.7} % Default value: 1
    	\centering
    	{\small
        	\caption{Distribution shift measured by Affinity score (\%). Lower the score, the transformation function triggers a larger distribution shift.}
    		\label{table_affinity}
    		\begin{tabular}{l  c}
    		    \hlineB{2.5}
    		    & Affinity \\
    		    \hline
        	    \multicolumn{1}{l|}{Rotation\cite{gidaris2018unsupervised}} & 58.06 \\
    		    \multicolumn{1}{l|}{LoRot-I} & 93.78 \\
    		    \multicolumn{1}{l|}{LoRot-E} & 90.15 \\
                \hlineB{2.5}
    		\end{tabular}
    	}
    \end{minipage}%\hfill
\end{table}

% \begin{wraptable}{hr}{0.16\textwidth}
% 	\vspace{-0.0cm}
% 	\hspace{-0.0cm}
% 	\centering
% 	{\small
% 		\begin{tabular}{l | c}
% 		    \hlineB{2.5}
% 		    \multicolumn{1}{c}{} & CIFAR10 \\
% 		    \hline
% 		    Baseline & 95.01 \\
% 		    +Rot (DA) & 92.76 \\
% 		    +Rot (MT) & 93.38 \\
% 		  %  +Rot (PT) & 95.68\% \\
		    
%             \hlineB{2.5}
% 		\end{tabular}
%     	\caption{CIFAR10 classification accuracy with rotation using different strategies.}
% 		\label{table_DAMTSS}
% 	}

% \end{wraptable}
% % \endgroup
% Self-supervision exploits data transformation to generate pretext tasks. 
% Indeed, proper data transformation can be also used as an augmentation technique to increase the amount and diversity of training samples which in turn, usually enhance the generalization capability of the model.
% \begin{wraptable}{tr}{4cm}

\begin{figure*}[t]
 % Default value: 6pt
    % \vspace{-0.3cm}
    % \vspace{-0.1cm}
    \begin{subfigure}{.24\textwidth}
      \centering
      % include first image
      \includegraphics[width=.8\linewidth]{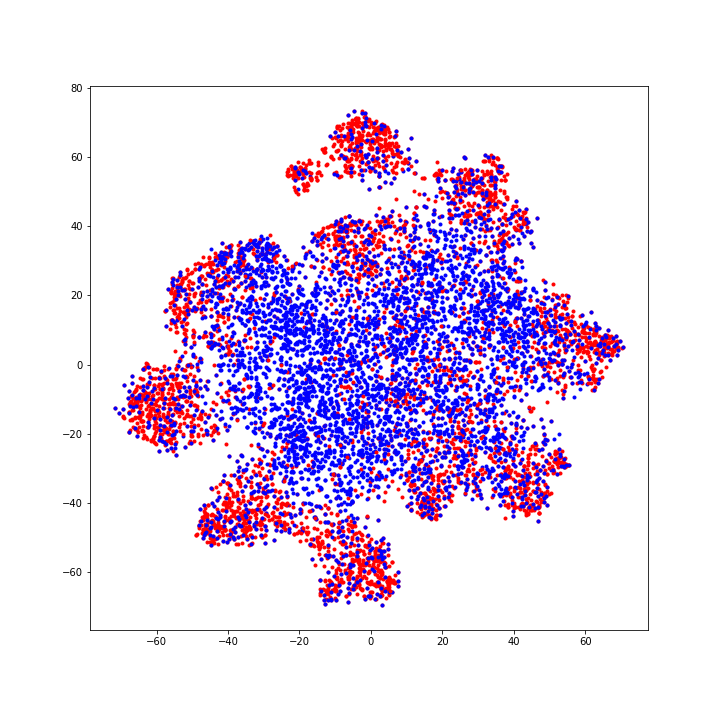}  
      \vspace{-0.25cm}
      \caption{Rotation}
      \label{fig:tsne-rot}
    \end{subfigure}\hspace{2mm}%\hfill% or  or \hspace{0.3\textwidth}
    \begin{subfigure}{.24\textwidth}
      \centering
      % include second image
      \includegraphics[width=.8\linewidth]{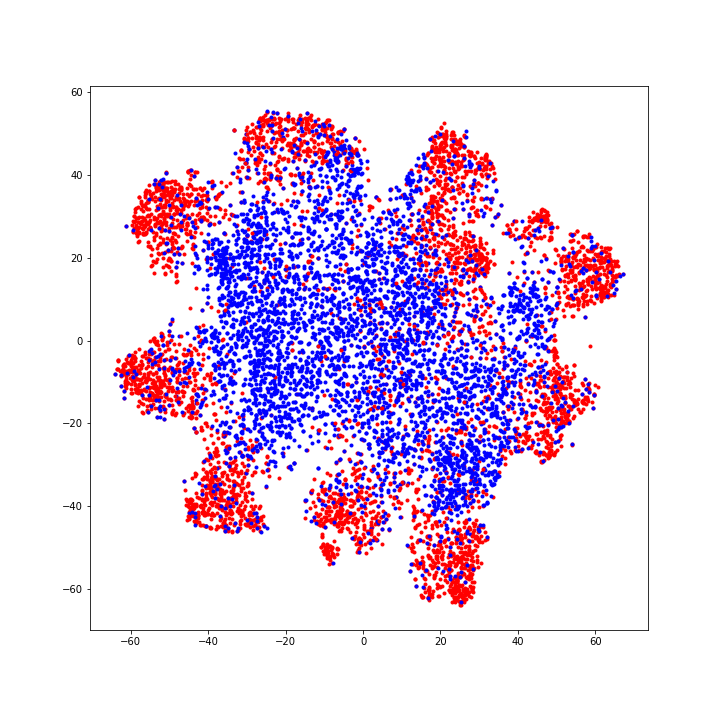}
      \vspace{-0.25cm}
      \caption{Jigsaw puzzle}
      \label{fig:tsne-jigsaw}
    \end{subfigure}\hspace{2mm}% or \hspace{5mm} or \hspace{0.3\textwidth}
    \begin{subfigure}{.24\textwidth}
      \centering
      % include second image
      \includegraphics[width=.8\linewidth]{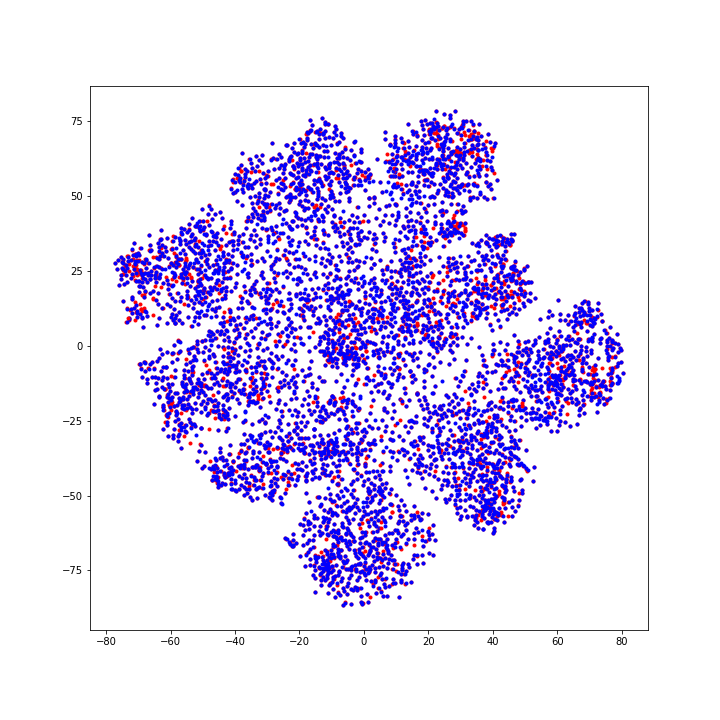} 
      \vspace{-0.25cm}
      \caption{LoRot-I}
      \label{fig:tsne-pr}
    \end{subfigure}
    \begin{subfigure}{.24\textwidth}
      \centering
      % include second image
      \includegraphics[width=.8\linewidth]{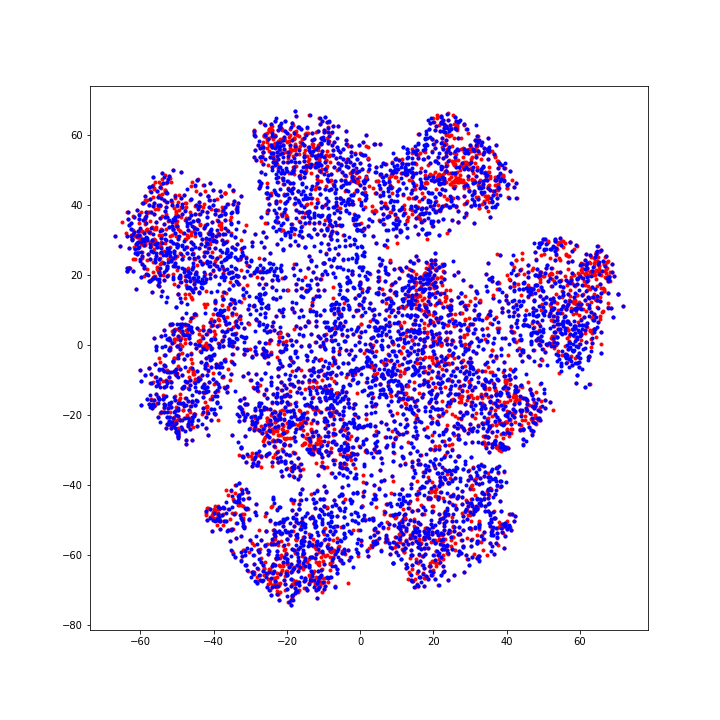} 
      \vspace{-0.25cm}
      \caption{LoRot-E}
      \label{fig:tsne-gr}
    \end{subfigure}
    \vspace{-0.2cm}
    %\caption{t-SNE~\cite{van2008visualizing} visualization of how each transformations maintain the data as it is from the viewpoint of the model. The model is trained by cross-entropy and tested on the normal and transformed test-set by each methods. Red dots are the original test samples and blue ones are the transformed test samples from CIFAR-10.}
    \caption{t-SNE~\cite{van2008visualizing} visualization of feature distributions of original (Red) and transformed (Blue) test samples to see the data distribution shifts induced by each data transformation deployed in different self-supervision tasks. Embedding features are extracted from the last convolution layer by forwarding original and transformed test samples to ResNet18~\cite{he2016deep} trained on the original train images of CIFAR-10.}
    \label{fig:tsne}
    \vspace{-0.4cm}
    % \captionsetup{belowskip=13pt}
    \setlength{\belowcaptionskip}{-10pt}
\end{figure*}

% 	\centering
% 	{\small
% 		\begin{tabular}{l | c}
% 		    \hlineB{2.5}
% 		    \multicolumn{1}{c}{} & Affinity \\
% 		    \hline
% 		    Rotation~\cite{gidaris2018unsupervised} & 58.06 \\
% 		    PatchRot-R & 93.78 \\
% 		    PatchRot-G & 90.15 \\
%             \hlineB{2.5}
% 		\end{tabular}
%     	\caption{Distribution shift measured by Affinity score. Higher score indicates reduced distribution-shift.}
% 		\label{table_affinity}
% 	}
% 	\vspace{-0.5cm}
% \end{wraptable}
Multi-task learning is an efficient and effective strategy when there exist multi objectives.
It only employs a single shared feature extractor and improves generalization by utilizing the domain-specific information contained in the training signals of related tasks~\cite{ruder2017overview, caruana1997multitask}.
% To employ such a strategy,  the transformation function should not incur the degree of the data distribution shift since the modified data distribution can impede the primary objective
To employ such a strategy, the transformation function should not incur the degree of the data distribution shift or should smooth the target label as the modified data distribution can impede the primary objective~\cite{zhang2017mixup, yun2019cutmix}. 
% To employ the multi-task learning strategy which is efficient when there exist multi objectives,
% the transformation function should not incur a degree of the data distribution shift or should smooth the label information as the modified distribution can impede the primary objective~\cite{zhang2017mixup, yun2019cutmix}. 
% To employ the multi-task learning strategy which is efficient when there exist multi objectives,
% the transformation function should not incur a degree of the data distribution shift or should smooth the label information as such transformation can impede the primary objective~\cite{zhang2017mixup, yun2019cutmix}. 
% \JH{should smooth the label information as such ~. This sentence is weird. So transformation should or should not smooth the label?}
However, previous self-supervision~\cite{gidaris2018unsupervised, noroozi2016unsupervised} and following works in supervised domains~\cite{hendrycks2019using, perera2020generative, carlucci2019domain, gidaris2019boosting} do not satisfy above which leads them to use inefficient ways to adopt in supervised domains (Discussed in next paragraph). 
% However, previous self-supervised works in superviesd domains~\cite{hendrycks2019using, perera2020generative, carlucci2019domain} neither satisfy the above. 
% Thus, we found the need to devise a new transformation that does not modify the semantics of the image.
% In our work, we focus only on satisfying the first condition: devising a transformation function for self-supervision that does not produce extremely different images.
% Since our goal is to develop a new self-supervision task to assist supervised learning, we devise a transformation that does not produce extremely different images.
% Since this paper discusses new self-supervision, we focus only on satisfying the first condition: devising a transformation function for self-supervision that does not produce extremely different images.
% produce extremely different images that do not correspond to the original distribution. 
Particularly, in Tab.~\ref{table_DAMTSS}, we conduct a simple experiment with the rotation task~\cite{gidaris2018unsupervised} to show how the classification result is affected when the transformed input is utilized to learn the primary task.
In the table, using the global rotation~\cite{gidaris2018unsupervised} only as an data augmentation technique (DA) shows the degraded performance.
Moreover, sharing the input features for multi-classifiers as the multi-task learning (MT) also provides worse performance than the supervised learning (Baseline). 
% Only the case of using a separate input (original image to primary classifier and transformed image to auxiliary classifier) denoted by PT, parallel-task learning, shows an improvement on the sacrifice of heavy cost.
To support our claims, we measure the distribution shift. 
% In Fig.~\ref{fig:tsne}, (a) and (b) indicate that transformations chosen for previous self-supervision generate distribution gap from its original distribution. 
To quantitatively measure the distribution shift, we use the affinity score~\cite{gontijo2020tradeoffs} on CIFAR-10 in Tab.~\ref{table_affinity}.
Affinity score is a metric to evaluate the distribution shift measured as follows: %that is calculated as \cref{equation_affinity}.
% Let $m$ be the model trained on train set and $A(m, D)$ be the model's accuracy when evaluated on dataset $D$. Further let $D_{val}$ be the validation set and $D^{\prime}_{val}$ be the augmented validation set from applying given stochastic augmentation strategy on $D_{val}$. Then, affinity score is measured as below:
\begin{equation}
% \label{eq_multitask}
% \vspace{-0.1cm}
    \text{Affinity} = \frac{A(m, D^{\prime}_{val})}{A(m, D_{val})},
    % A(m, D^{\prime}_{val}) / A(m, D_{val})
    \label{equation_affinity}
    % \vspace{-0.1cm}
\end{equation}
where $m$ is a model trained on training set, and $A(m, D)$ is the accuracy of $m$ on a dataset $D$. $D_{val}$ and $D^{\prime}_{val}$ are the original and augmented validation sets, respectively.
Furthermore, we also qualitatively show in Fig.~\ref{fig:tsne} (a) and (b) that transformations from previous self-supervision lead distribution gap from its original one.
Therefore, we can derive that the data distribution shift triggers unstable training and a new transformation that maintains the semantics of the image is needed to employ multi-task learning.
% Thus, we found the need to devise a new transformation that does not modify the semantics of the image.
% Since our goal is to develop a new self-supervision task to assist supervised learning, we devise a transformation that does not produce extremely different images.
% Fig.~\ref{fig:tsne} is discussed further in Sec.~\ref{subsection:methodology}.

Applicability is an another important aspect of an auxiliary task in practical view. Devising lightweight architectures and methods is one of the current research trend~\cite{khan2020survey}. However, since current self-supervised methods are not studied thoroughly in supervised learning, they often trigger high extra cost. 
Specifically, previous works~\cite{perera2020generative, hendrycks2019using, carlucci2019domain} adopted self-supervision into the supervised domains in rather inefficient ways: parallel-task learning strategy or label augmentation strategy.
We define parallel-task learning as each separate input sets being forwarded to handle each tasks in contrast to multi-task learning. 
For more details, See Fig.~\ref{figure_CAM} (c).
We further note that label augmentation requires all possible transformations to be applied per sample at both the training and inference times.
At this point, when the usage of pretext tasks in the supervised domain is increasing on the sacrifice of expensive costs, applicability is the next mission for self-supervision to be more widely applied in supervised domains.

\subsection{Localizable Rotation (LoRot)}
\label{subsection:methodology}
\begin{figure*}[t]
 % Default value: 6pt
%  \vspace{-0.2cm}
\centering
    \begin{subfigure}{0.45\textwidth}
      \centering
      \includegraphics[width=1\linewidth]{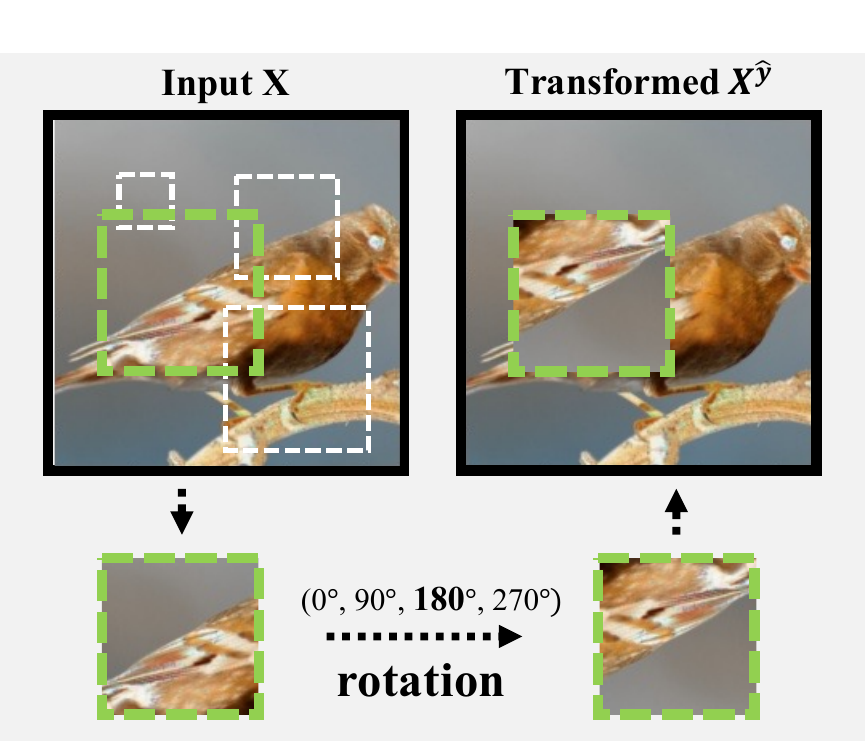}  
    %   \vspace{-0.3cm}
      \caption{LoRot-I}
    \end{subfigure}\hspace{2mm}%\hfill% or  or \hspace{0.3\textwidth}
    \begin{subfigure}{0.45\textwidth}
      \centering
      \includegraphics[width=1\linewidth]{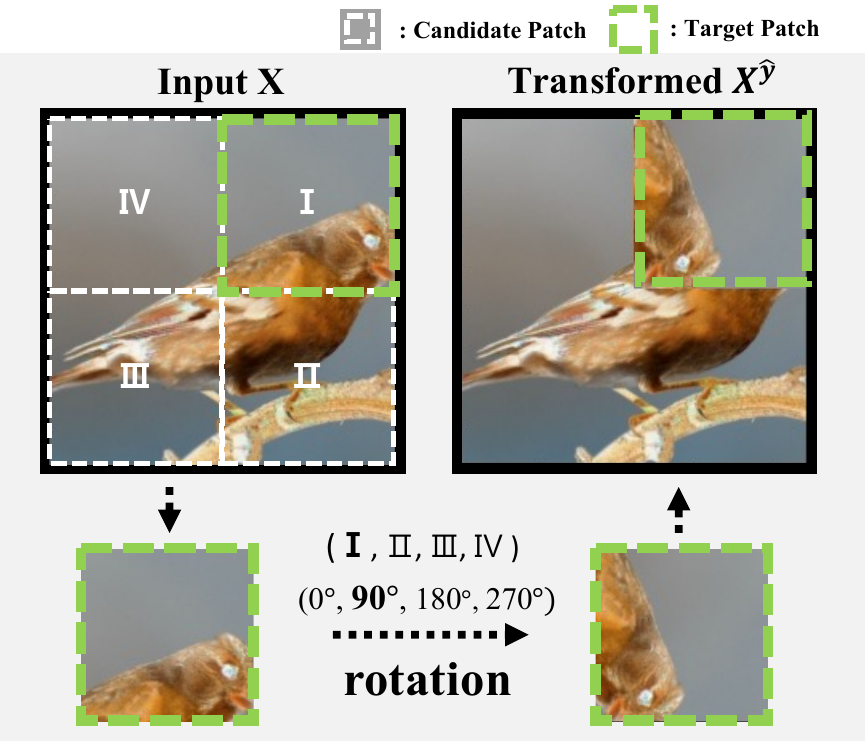} 
    %   \vspace{-0.3cm}
      \caption{LoRot-E}
    %   \label{fig:PatchRot-G}
    \end{subfigure}\hspace{2mm}% or \hspace{5mm} or \hspace{0.3\textwidth}

    % \vspace{-0.2cm}
    %\caption{Difference between PatchRot and PatchRot-Grid. Both (a) and (b) show possible transformations for each method. White-thin dotted line indicates possible candidates for patch while Green-think dotted line is the selected patch in each method.}
    \caption{Illustration of LoRot-I and LoRot-E. (a) LoRot-I draws and rotates a random patch from the image, while (b) LoRot-E chooses and rotates a cell from the predefined grid layout. For both methods, the degree of rotation is randomly chosen from $\{0^{\circ},90^{\circ},180^{\circ},270^{\circ}\}$. Note that, white and green boxes indicate possible and selected patches in this example, respectively.}
    \label{Fig1_DifferencebetweenPRandPRG}
    \vspace{-0.3cm}
    % \captionsetup{belowskip=13pt}
    \setlength{\belowcaptionskip}{-10pt}
\end{figure*}

Localizable rotation is designed to rotate a local region.
To solve the localizable rotation task, the model should first localize the patch and then identify the high-level clues to predict the rotation degree, e.g., object parts such as eyes, tails, and heads within the patch\cite{gidaris2018unsupervised}.
Therefore, an explicit localization task to predict the position of the patch may not be necessary to guide the model to learn the localization capability.
In this context, we introduce two versions of LoRot with explicit and implicit localization tasks as shown in Fig.~\ref{Fig1_DifferencebetweenPRandPRG}.
For the rest of this paper, we define LoRot-E and LoRot-I as each LoRot having the localization task explicitly and implicitly.
% Localizable rotation is designed to rotate a local region. 
% To solve the localizable rotation task, the model should first localize the patch and then identify the high-level clues, e.g., object parts such as eyes, tails, and heads within the patch\cite{gidaris2018unsupervised}.
% Therefore, an explicit localization task to predict the position of the patch may not be necessary to enforce the model to learn the localization capability.
% We introduce two versions of LoRot with explicit and implicit localization tasks as shown in \cref{Fig1_DifferencebetweenPRandPRG}.
% For the rest of this paper, we define LoRot-E and LoRot-I as each LoRot having the localization task explicitly and implicitly.

% It drives the model to locate the patch and localize the rotated salient objects~\cite{gidaris2018unsupervised}. 
% Due to its property of forcing to learn spatial information, we thought explicit localization task is not necessary. 
% Therefore, we introduce two versions of LoRot each with explicit and implicit localization tasks. For the rest of this paper, we define LoRot-E and LoRot-I as each LoRot having the localization task explicitly and implicitly.
% For LoRot, localization task can be given both explicitly and implicitly. Explicit task can be provided by pre-defining the position to generate pretext labels. On the other hand, without the localization task, the model is still indirectly forced to learn the spatial information since the rotation task requires the location of rotated salient objects~\cite{gidaris2018unsupervised}. For the rest of this paper, we define LoRot-E and LoRot-I as LoRot having the localization task explicitly and implicitly, each.

Let $X \in \mathbb{R}^{H \times W \times C}$ be a training image with the width $W$, height $H$, and channels $C$, and $y$ be its class label of supervised learning.
Unless mentioned, LoRot is used in the form of multi-task learning with two classifiers each for the primary task and localizable rotation.
Let also the feature extractor and two softmax classifiers be ${F}_{\theta}, \sigma_u$, and $\sigma_v$ parameterized by $\theta$, $u$ and $v$, respectively. We also define the transformation function $T$ and the pretext label $\hat{y}$ in which $T$ generates the transformed sample $X^{\hat{y}}$ as follows:
\begin{equation}
% \label{eq_multitask}
% \vspace{-0.1cm}
    X^{\hat{y}} = T(X|\hat{y}, S),
    % \vspace{-0.1cm}
\end{equation}
where $S$ stands for patch selection strategy. 
We define possible rotation degrees to (${0, 90, 180, 270^\circ}$) following the rotation task~\cite{gidaris2018unsupervised}. 
% We define possible rotation degrees to (${0^\circ, 90^\circ, 180^\circ, 270^\circ}$) following the rotation task~\cite{gidaris2018unsupervised}. 
% We follow the global rotation~\cite{gidaris2018unsupervised} that possible rotation degrees are (${0^\circ, 90^\circ, 180^\circ, 270^\circ}$). 
Then, as shown in Fig.~\ref{Fig1_DifferencebetweenPRandPRG}, the number of classes for LoRot-I would be 4 and 16 for LoRot-E with the position in the 2x2 grid layout. Note that, for LoRot-E, we keep redundant cases with $0^\circ$ at every cell as we pursue to place more weights on the original image.

Moreover, we define $P_{u}(X^{\hat{y}}) = \sigma_u(F_\theta(X^{\hat{y}}))$ and $P_{v}(X^{\hat{y}}) = \sigma_v(F_\theta(X^{\hat{y}}))$ as the probability distributions over the labels of the primary and pretext tasks, respectively.
When $P^{j}(.)$ is the probability of the j-th class with a batch of $N$ training images $\{X_{i}\}_{i=1}^{N}$, the overall objective is: 
% \vspace{-0.2cm}
\begin{equation}
\label{eq_multitask}
    % \vspace{-0.15cm}
    \min_{\theta}-\frac{1}{N}\sum_{i=1}^{N}(\log(P_{u}^{y}(X_{i}^{\hat{y}})) +\lambda \log(P_{v}^{\hat{y}}(X_{i}^{\hat{y}}))),
    % \vspace{-0.15cm}
\end{equation}
where $\lambda$ is a hyperparameter to control the weight of learning LoRot.

\textbf{Patch Selection.}
We use different strategies to generate patches for LoRot-E and LoRot-I. Simply put, we pre-define the 2x2 grid layout for LoRot-E to easily design the localization task. Then, the sampling method S does not need any parameters. Specifically, we divide each image into a $K \times K$ uniform grid and rotate a single cell of the grid with the dimension of $\mathbb{R}^{\frac{H}{K} \times \frac{W}{K} \times C}$. 
% In this paper, we set K to 2 which each quadrant of image can be the victim.
In this paper, we set $K$ to 2 which each quadrant of an image can be the target.

Meanwhile, we choose random sampling method for LoRot-I. We randomly sample a length $l$ and the position of the top-left corner $(p_{x}, p_{y})$ from the uniform distribution \textrm{U} to form sampling strategy $S$ as follows:
% \vspace{-0.2cm}
\begin{equation}
% \[
% \textbf{if}\: \text{LoRot-E}, \: l = \textrm{min}(W/2, H/2), \:
% \textbf{elif}\: \text{LoRot-I}
S(l, p_{x}, p_{y})\: 
\begin{cases}
    \:l \sim \textrm{U}(2, \textrm{min}(\lfloor W/2\rfloor, \lfloor H/2\rfloor)),\\[-2pt]
    \:p_{x} \sim \textrm{U}(0, W-l),\\[-2pt]
    \:p_{y} \sim \textrm{U}(0, H-l)
\end{cases}
% \vspace{-0.2cm}
% \]
\end{equation}

% \begin{equation}
% \label{eq_sampling}
% % \text{\~}
%     l \sim \textrm{U}(2, \textrm{min}(\lfloor W/2\rfloor, \lfloor H/2\rfloor)),\; p_{x} \sim \textrm{U}(0, W-l),\; p_{y} \sim \textrm{U}(0, H-l).
% \end{equation}
Note that only a square-shaped patch is used in our work for simplicity. Also, we limit the length of the patch up to half of $\min(H, W)$ to prevent rotating an overly large region. Next, we detail how LoRot satisfies the desired properties.

% \begin{figure*}
%     \centering
%     \includegraphics[scale=0.46]{fig/Final_cam+MT.pdf}

%     \caption{(a) Comparison of class activation maps (CAMs)~\cite{zhou2016learning} of differently learned models. Rot indicates the global rotation task. DA, PT and MT stand for (refer Sec.~\ref{subsection:discussion}) the strategies of utilizing the rotation task, Data Augmentation, Parallel-Task learning, and Multi-Task learning, as illustrated in (b), (c) and (d), respectively. PatchRot and PatchRot-G are applied to the baseline by MT as designed. The CAMs show that our PatchRot (and -G) provides the activation of higher coverage to the object compared to global rotations. In other words, PatchRot auxiliary task encourages the model to learn rich features. Best viewed in color.}
%     \label{figure_CAM}
%     % \vspace{-1.0cm}
% \end{figure*}

\begin{figure*}[t]
    \centering
    % \vspace{-0.1cm}
    \includegraphics[scale=0.39]{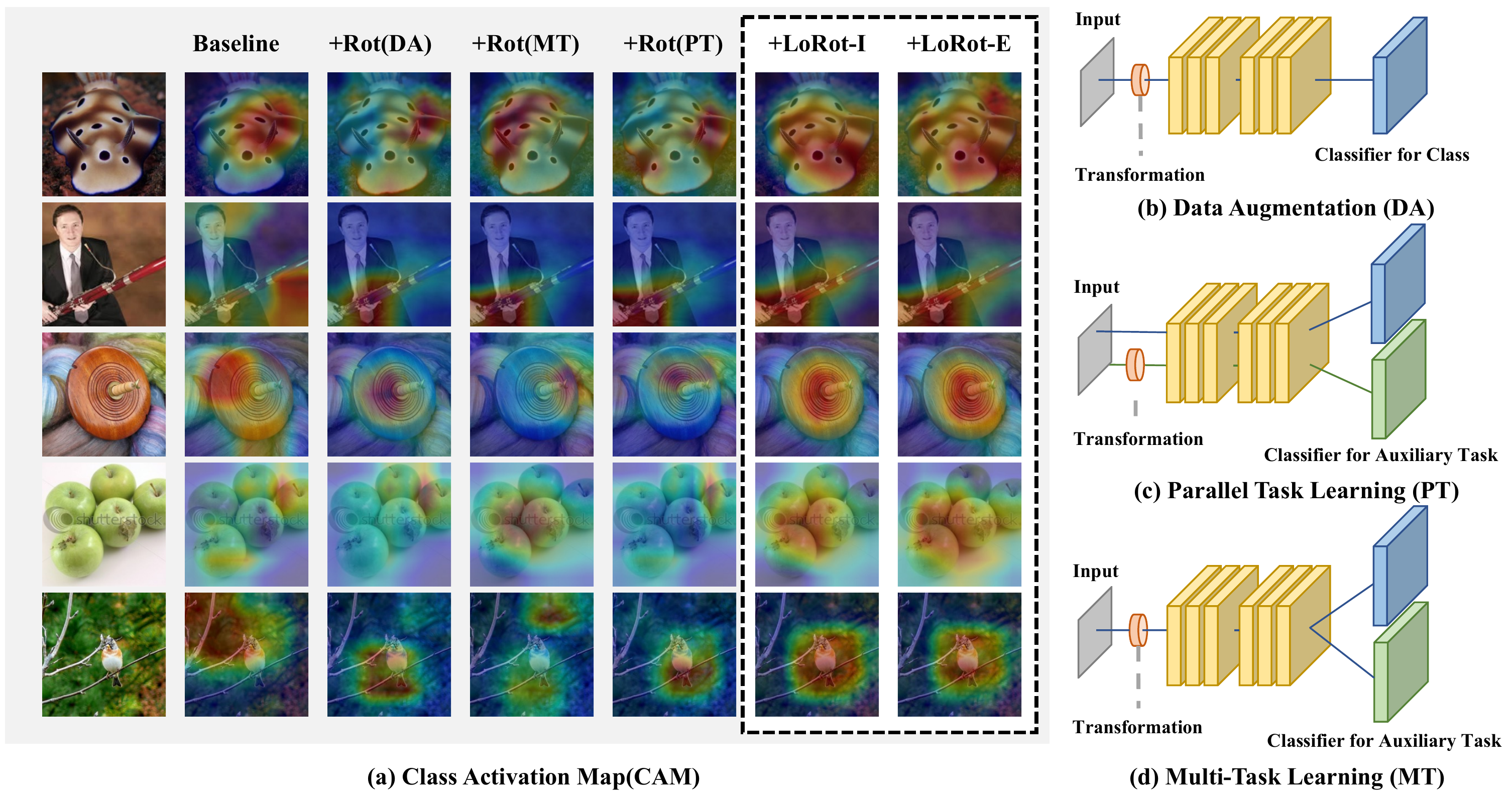}
    \vspace{-0.1cm}
    \caption{(a) Comparison of class activation maps (CAMs)~\cite{zhou2016learning} of differently learned models. Rot indicates the global rotation task. DA, PT and MT stand for each strategy of utilizing the rotation task: Data Augmentation, Parallel-Task learning, and Multi-Task learning, as illustrated in (b), (c) and (d). DA and MT take augmented input to predict single- or multi-tasks. On the other hand, PT requires separate input batches to predict primary and auxiliary tasks, respectively. LoRot-I and LoRot-E are applied to the baseline by MT as designed. The CAMs show that our LoRot provide the activation of higher coverage to the object compared to global rotations. In other words, LoRot auxiliary task encourages the model to learn rich features. Best viewed in color.}
    \label{figure_CAM}
    \vspace{-0.3cm}
\end{figure*}
\textbf{Rich Representations.}
LoRot encourages the model to consider even the less-discriminative features by setting rotation prediction quizzes on different locations within an image. Particularly, the model should learn rich features to solve rotation tasks for patches of various sizes at different locations. One may ask that the LoRot can produce many useless features. For instance, the rotation problem with patches totally outside object regions can disturb learning good representations. However, this is alleviated by joint optimization with the supervised objective. Since the primary loss is rather dominant compared to LoRot's, such certainly unnecessary features are dropped throughout the training iterations. 
To support our argument, we investigate the class activation maps\cite{zhou2016learning} of various models trained with only the primary task (Baseline), and the primary task with auxiliary tasks of global rotation (Rot(*)), and LoRot. 
\begin{wrapfigure}{t!}{5.5cm}
\centering
    \vspace{-0.1cm}
    \includegraphics[width=5.5cm]{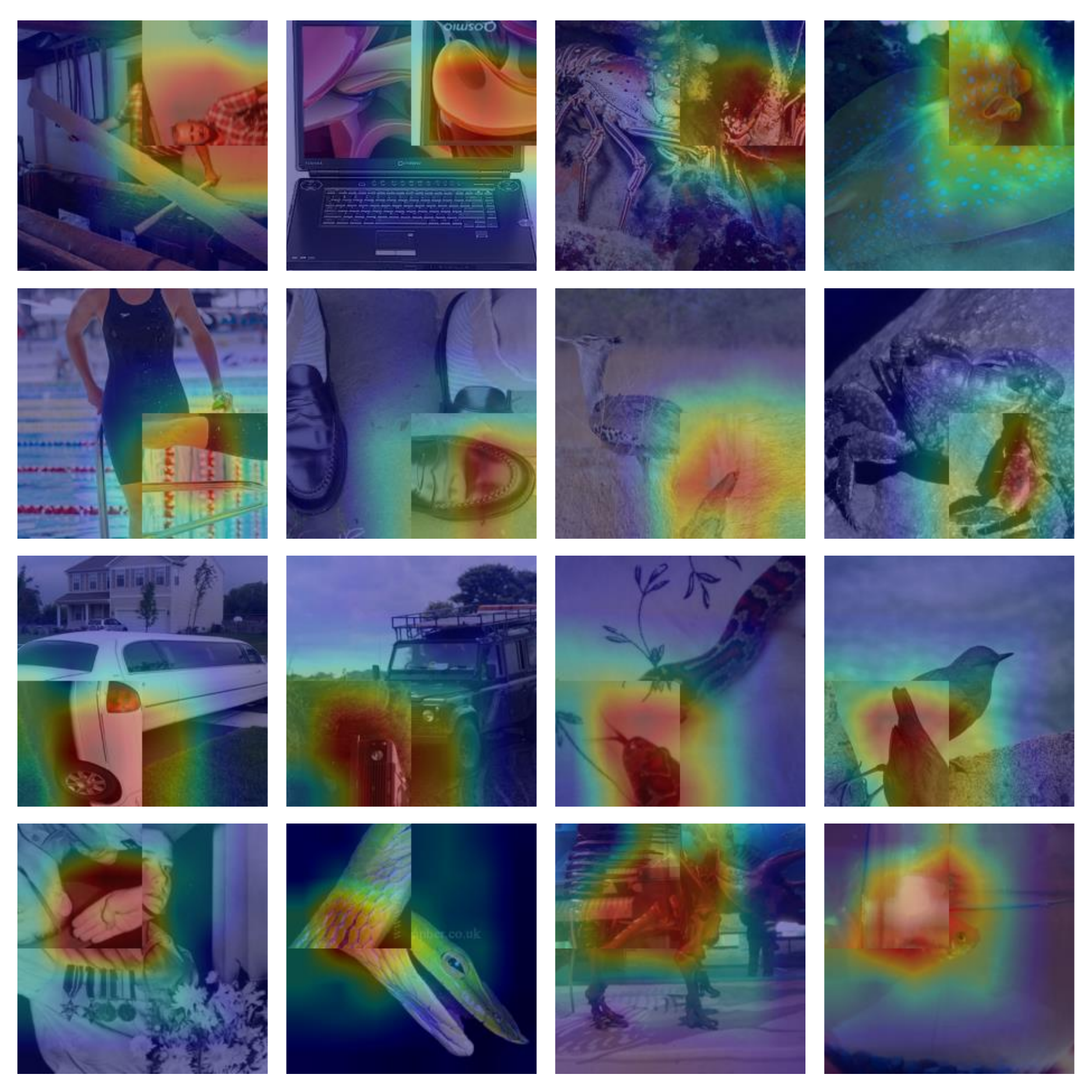}
    % \includegraphics[scale=0.3]{fig/CAMforAuxtask.pdf}
    % \includegraphics[scale=0.4]{fig/CAMforAuxtask.pdf}
    % \vspace{-0.2cm}
    \caption{Class activation mapping visualizations for predicting LoRot. From top to bottom, we show that the model focuses on the high-level clues in the rotated patch at each quadrant, i.e., in the first column, the model spotlights the head, leg, wheel, and hand.}
    \label{fig:CAMAuxtask}
    \vspace{-0.6cm}
\end{wrapfigure} 
As shown in Fig.~\ref{figure_CAM} (a), LoRot provides larger coverage of activations on the object compared to others. We also quantitatively validate the aforementioned in terms of model robustness and localization performance in Sec.~\ref{section:Experiments}.
% \begin{figure}[t]
%     \centering
%     % \includegraphics[scale=0.4]{latex/fig/CAMforAuxtask.pdf}
%     \includegraphics[scale=0.38]{fig/CAM_AccuracyCurve_Auxtask.pdf}
%     \caption{(Left) Class activation mapping visualizations for predicting LoRot. From top to bottom, we show that the model focus on the high-level clues in the rotated patch at each quadrant, i.e., in the first column, the model spotlights the head, leg, wheel, and hand. (Right) Accuracy curves of LoRot-E, LoRot-I, and rotation task during training (Y-axis : Accuracy, X-axis : Epoch). Due to different strategies applied between LoRot and rotation in supervised setting, we only illustrate accuracy curves of LoRot variants.
%     }
%     \label{fig:CAMAuxtask}
% \end{figure}
% \begin{figure}[t]
%     \centering
%     % \vspace{-0.1cm}
%     \includegraphics[scale=0.3]{fig/CAMforAuxtask.pdf}
%     % \includegraphics[scale=0.4]{fig/CAMforAuxtask.pdf}
%     % \vspace{-0.2cm}
%     \caption{Class activation mapping visualizations for predicting LoRot. From top to bottom, we show that the model focus on the high-level clues in the rotated patch at each quadrant, i.e., in the first column, the model spotlights the head, leg, wheel, and hand.}
%     \label{fig:CAMAuxtask}
%     % \vspace{-0.6cm}
% \end{figure}
As discussed, LoRot discovers features and asks their necessity to the primary task. 
In Fig.~\ref{fig:CAMAuxtask}, we show how LoRot discovers sub-discriminative features.
Whenever the position of the patch moves, the model adjusts its focus to the other parts of the image to solve the auxiliary task.
This also promotes the model to neglect unnecessary parts to solve specific LoRot tasks since the salient clues for predicting LoRot is random.
% On top of this, we also observe that the boundaries of the rotated patch are not the shortcut in LoRot since boundaries do not contain informative features for predicting rotation degree.
% In addition, we plot accuracy curves for LoRot in Fig.~\ref{fig:CAMAuxtask} (Right). This shows that both LoRot tasks are learnable, and LoRot-I is a harder task than the 

% \cref{fig:CAMAuxtask} shows that whenever the location of the patch moves, the model adjusts its focus to the other parts of the image to solve the auxiliary task. Therefore, since the salient part for predicting LoRot is random during training and the scaling factor $\lambda$ of the loss of auxiliary tasks is much smaller than the primary one, the model neglects unnecessary parts that were only needed to solve specific auxiliary tasks.
% \WJ{Intuitively}, it can be explainable that the LoRot discovers features and asks their necessity to the primary task. \cref{fig:CAMAuxtask} shows that whenever the location of the patch moves, the model adjusts its focus to the other parts of the image to solve the auxiliary task. Therefore, since the salient part for predicting LoRot is random during training and the scaling factor $\lambda$ of the loss of auxiliary tasks is much smaller than the primary one, the model neglects unnecessary parts that were only needed to solve specific auxiliary tasks.

% \begin{wrapfigure}{r}{0.45\textwidth

\textbf{Maintaining Data Distribution.}
\label{avoiding_data_distribution_shift}
Unlike the existing transformations of pretext tasks, LoRot is less likely to incur data distribution shift since it only carries out geometric transformations locally.
Specifically, most parts are kept intact in LoRot so that data distribution shift is restrained. Not just the smaller number of the transformed pixels but also the preserved high-level semantics contribute to keep the distribution close to the original one.
In Fig.~\ref{fig:tsne} (c) and (d), we observe that LoRot maintains the data distribution as the blue obscures the red in the embedding space. 
Affinity score also validates it as shown in Tab.~\ref{table_affinity}.

\textbf{Applicability.}
To apply self-supervision of LoRot, we adopt multi-task learning so that a single transformed input is shared by both the primary and pretext tasks as shown in Fig.~\ref{figure_CAM} (d).
This provides the advantages in the computational cost and easy deployment to existing models since LoRot only requires one extra classifier without requiring multi-batches.
% increasing the number of samples forwarded. 
In fact, previous pretext tasks usually require several times more samples to be forwarded and backpropagated to achieve their performances as they apply all possible transformations per sample~\cite{hendrycks2019using, lee2020self}.
% Throughout Sec.~\ref{section:Experiments}, we validate LoRot's high applicability in terms of lightweight tasks and boosting performance.
Throughout Sec.~\ref{section:Experiments}, we validate LoRot's high applicability, a lightweight task boosting baselines' performances.
Particularly, we observe that LoRot is not only complementary to standard baselines, but also to contrastive approaches in Tab.~\ref{table_OOD} and Tab.~\ref{table_Sup_CIFAR_complementary}.
This spotlights LoRot as an easily attachable self-supervised module for many supervised methods.

% Furthermore, as we observe LoRot is complementary to contrastive learning in Sec.~\ref{}, LoRot is an easily attachable module for many fully supervised methods.
% We validate its high applicability in Sec.~\ref{section:Experiments} in terms of lightweight tasks and boosting performance.

%% file: _4_experiments.tex
\begingroup
\setlength{\tabcolsep}{3.7pt} % Default value: 6pt
\renewcommand{\arraystretch}{1.1} % Default value: 1
\begin{table*}[t!]
	\centering
	\caption{AUROC scores for distinguishing in- and out-distribution data for image classification. The model is trained with CIFAR-10 dataset and evaluated on both CIFAR-10 and each OOD dataset. `$\ast$' indicates our reproduced version based on official implementations to unify the backbone network and training protocols. All experiments are averaged over five runs and `$\pm$' denotes the standard deviation. `FS' and `IN' stand for the number of forwarded samples to train each model and ImageNet, respectively. 
% 	Bolds represent the highest auroc score and the underlines indicate the second best one.
    }
    	\label{table_OOD}
	{\scriptsize
% 	{\footnotesize
% 	{\small
		\begin{tabular}{l| c c c c c c | c}
		    \hlineB{2.5}
		  %  \multicolumn{5}{r}{In Distribution : CIFAR10} \\
            % \cmidrule(r){2-8}
            % \multicolumn{8}{c}{Out Distribution} \\ 
            Method & SVHN & LSUN & IN & LSUN (FIX) & IN (FIX) & CIFAR-100 & FS \\
			\hlineB{2.5}
            Cross Entropy$\ast$ & 84.6$_{\pm5.2}$ & 90.9$_{\pm0.7}$ & 87.8$_{\pm1.4}$ & 84.3$_{\pm1.0}$ & 85.3$_{\pm0.6}$ & 83.5$_{\pm0.5}$ & 5M\\
 			Cutmix~\cite{yun2019cutmix}$\ast$ & 75.5$_{\pm9.5}$ & 92.5$_{\pm3.3}$ & 92.1$_{\pm2.0}$ & 86.2$_{\pm1.0}$ & 84.3$_{\pm1.0}$ & 80.9$_{\pm1.1}$ & 5M\\
 			SLA+SD~\cite{lee2020self}$\ast$ & 89.1$_{\pm4.4}$ & 90.7$_{\pm1.3}$ & 89.8$_{\pm0.8}$ & 82.9$_{\pm1.6}$ & 86.0$_{\pm0.7}$ & 83.6$_{\pm0.4}$ & 20M \\
			Rotations~\cite{hendrycks2019using}$\ast$ & 96.1$_{\pm1.8}$ & 97.3$_{\pm0.5}$ & 96.9$_{\pm0.9}$ & 91.0$_{\pm0.4}$ & 91.8$_{\pm0.2}$ & 89.1$_{\pm0.4}$ & 25M \\
% 			APR-SP~\cite{}& \textbf{97.7}$_{\pm NA}$ & 97.9$_{\pm NA}$ & 96.3$_{\pm NA}$ & 93.7$_{\pm NA}$ & 92.8$_{\pm NA}$ & 89.5$_{\pm NA}$ & - \\
            \hline
     		SupCLR~\cite{khosla2020supervised} & 97.3$_{\pm 0.1}$ & 92.8$_{\pm 0.5}$ & 91.4$_{\pm 1.2}$ & 91.6$_{\pm 1.5}$ & 90.5$_{\pm 0.5}$ & 88.6$_{\pm 0.2}$ & 70M \\
 			CSI~\cite{tack2020csi} & 96.5$_{\pm 0.2}$ & 96.3$_{\pm 0.5}$ & 96.2$_{\pm 0.4}$ & 92.1$_{\pm 0.5}$ & 92.4$_{\pm 0.0}$ & 90.5$_{\pm 0.1}$ & 280M \\
 			\hline
 			LoRot-I  & 92.6$_{\pm2.1}$ & \underline{98.6}$_{\pm0.7}$ & \underline{98.0}$_{\pm0.8}$ & 94.4$_{\pm0.9}$ & 93.6$_{\pm1.0}$ & 90.1$_{\pm0.7}$ & 5M \\
            LoRot-E  & 94.4$_{\pm0.9}$ & \textbf{98.7}$_{\pm0.6}$ & \textbf{98.1}$_{\pm0.5}$ & 94.1$_{\pm0.3}$ & 93.1$_{\pm0.4}$ & \underline{90.6}$_{\pm0.3}$ & 5M \\
            \hline
 			CSI+LoRot-I & \textbf{97.7}$_{\pm0.6}$ & 98.3$_{\pm0.1}$ & \underline{98.0}$_{\pm0.3}$ & \textbf{95.7}$_{\pm0.1}$ & \textbf{95.6}$_{\pm0.1}$ & \textbf{93.8}$_{\pm0.0}$ & 280M \\
            CSI+LoRot-E & \underline{97.5}$_{\pm0.4}$ & 98.0$_{\pm0.2}$ & 97.8$_{\pm0.1}$ & \underline{95.5}$_{\pm0.2}$ & \underline{95.4}$_{\pm0.2}$ & \textbf{93.8}$_{\pm0.1}$ & 280M \\
            \hlineB{2.5}
        	
		\end{tabular}
	}
\end{table*}
% \endgroup
\section{Experiments}
\label{section:Experiments}
% Throughout this section, we validate PatchRot in various supervised settings in auxiliary manner. 
%In section 4.1, we evaluate classification accuracy and conduct Out-of-Distribution Detection, Open-set Recognition and Imbalanced classification experiments in each section~\ref{subsection:OSRExperiments}, \ref{subsection:OODExperiments}, \ref{subsection:ImbExperiments}. Unless otherwise mentioned, we multitask learning with additional classifier when applied to other methods including fully supervised method.
% We first examine the robustness of LoRot on OOD detection, imbalanced classification, and adversarial attack in Sec.~\ref{subsection:Robustness}.
% Afterwards, we validate the generalization capability on image classification, weakly supervised object localization (WSOL), and transfer learning in Sec.~\ref{subsection:Classification}. 
We first examine the robustness of LoRot in Sec.~\ref{subsection:Robustness} and validate the generalization capability in Sec.~\ref{subsection:Classification}. 
Unless otherwise mentioned, LoRot is applied to supervised baseline with cross-entropy loss. 
For baselines, we compare rotation~\cite{gidaris2018unsupervised}, the most popular pretext task in supervise domain, and previous works that adopted self-supervision in supervised domains (Rotations~\cite{hendrycks2019using}, SLA+SD~\cite{lee2020self}, and SSP~\cite{yang2020rethinking}).
We also compare SOTA methods between benchmarks and show that contrastive learning is a complementary method, not our baseline.
For other pretext tasks, we claim these are neither our baseline nor better than our baselines since our baselines are modified versions for supervised domain on top of existing pretext tasks.
Throughout this section, we use bolds and underlines to represent the best and the second best scores.
Furthermore, as LoRot is robust to $\lambda$, we set $\lambda$ to 0.1 for all experiments and further explore the effects of $\lambda$ in the supplementary.
% Code and models will be publicized to support future studies.

\subsection{Robustness}
\label{subsection:Robustness}
% In this section, we conduct experiments on three challenging robustness tasks; out-of-distribution detection, imbalanced classification, and adversarial attack.

% Note that LoRot generally work well on many benchmarks, so that tuning the hyperparameter per benchmark is not necessary. In this section, we set the only hyperparameter $\lambda$ to 0.1 for all three experiments. 
% We further explore the effects of $\lambda$ in details in Appendix.

\subsubsection{Out-Of-Distribution Detection}
\label{subsection:OODExperiments}
 is to assess the model's uncertainty against unknown data. 
It is essential when deploying the model in real-world systems since DNNs are vulnerable to shortcut learning~\cite{geirhos2020shortcut, nguyen2015deep, hendrycks2016baseline}.
%  is to assess the model's understanding of given samples~\cite{liang2017enhancing}.
% It is particularly important since DNNs are vulnerable to shortcut learning~\cite{geirhos2020shortcut, nguyen2015deep, hendrycks2016baseline}.
% We state that OOD detection is a particularly important benchmark to assess the model's understanding of given samples due to the vulnerability of DNN to shortcut learning. 
For the experiment, we train the model with CIFAR-10 ~\cite{krizhevsky2009learning} which we call it in-distribution dataset. 
Then, we use SVHN~\cite{netzer2011reading}, resized ImageNet and LSUN~\cite{liang2017enhancing}, fixed versions of ImageNet and LSUN~\cite{tack2020csi}, and CIFAR-100~\cite{krizhevsky2009learning} as out-of-distribution datasets.
We compare our method against the previous SOTA works on OOD detection~\cite{hendrycks2019using, tack2020csi} as well as approaches that utilize a rotation technique~\cite{lee2020self} and regional modification~\cite{yun2019cutmix} to enhance the robustness.
% Specifically, Rotations~\cite{hendrycks2019using} uses rotation and affine transformation to form the pretext task.
% SLA+SD~\cite{lee2020self} augments label information with rotation. 
% Moreover, CSI~\cite{tack2020csi} deploys rotation as shifting transformation in contrastive learning. 
Unlike the previous SOTA works that require huge costs either at the training~\cite{tack2020csi} or inference time~\cite{hendrycks2019using} to yield their best performance, LoRot does not require huge costs neither at the training nor inference time.
Indeed, we only take 3.6\% of training time compared to \cite{tack2020csi} and 50\% of inference time compared to \cite{hendrycks2019using} (Measured with 2 Quadro RTX 8000).
Still, we acquired significant gains on detecting OOD samples. For fair comparison, we use the ResNet18~\cite{he2016deep} following the previous SOTA work~\cite{tack2020csi} to unify the benchmarks.

As we can see in Tab.~\ref{table_OOD}, our proposed LoRot outperforms the state-of-the-art methods on five benchmarks. 
To measure the performance of LoRot, we utilize the KL-divergence between the softmax predictions and the uniform distribution as in \cite{hendrycks2018deep, hendrycks2019using}.
However, we use the softmax predictions for SLA+SD~\cite{lee2020self} and CutMix~\cite{yun2019cutmix} as the softmax results fit better with their methods. 
The results for SupCLR~\cite{khosla2020supervised} and CSI~\cite{tack2020csi} are from its paper and we further report the performances of LoRot when applied to contrastive approach, CSI.
% As we report the performances for SupCLR~\cite{khosla2020supervised} and CSI~\cite{tack2020csi} from its original paper, we employ softmax predictions for CSI+LoRot following the protocol from CSI~\cite{tack2020csi}.
Interestingly, we observe that the AUROC score of CutMix~\cite{yun2019cutmix} degrades on harder benchmarks in OOD detection.
We conjecture that the label smoothing effect of CutMix could degrade their robustness to unseen samples in harder benchmarks.
In contrast, LoRot consistently improves the baselines by large margin
(including 11\%p and 4\%p improvement on LSUN(FIX) dataset to cross-entropy and CSI, respectively).
For the slightly low performance on SVHN, we think that it is because there is no difference when a small patch is rotated against a plain background of the SVHN dataset.
Thus, we believe LoRot-E is better when images are composed of a simple background and, otherwise both approaches would work fine.
\begingroup
\setlength{\tabcolsep}{7.0pt} % Default value: 6pt
\renewcommand{\arraystretch}{0.8} % Default value: 1
\begin{table*}[t]
    \centering
    {
        \caption{Imbalanced classification accuracy (\%) on CIFAR-10/100. We add LoRot and other self-supervised approaches on LDAM-DRW and compare the gains. 
    % Bolds and underlines indicate the highest and the second highest auroc scores.
    }
    \label{table_IM0.01}
    % \scriptsize
    \small
        \begin{tabular}{l | c c c | c c c}
            \hlineB{2.5}
          %  \multicolumn{5}{|c|}{\textbf{Review of learning needs}} & \multicolumn{4}{c|}{\textbf{Development plan}}\\ \hline
            % \multicolumn{1}{c|}{} & \multicolumn{3}{c|}{Imbalanced CIFAR10} & \multicolumn{3}{c}{Imbalanced CIFAR100} \\
            % \cmidrule(r){1-4} \cmidrule(r){4-7}
            Imbalance Ratio & 0.01 & 0.02 & 0.05 & 0.01 & 0.02 & 0.05  \\
            \hlineB{2.5}
        
            LDAM-DRW~\cite{cao2019learning} & 77.03 & 80.94 & 85.46 & 42.04 & 46.15 & 53.25\\
            \hline
            + Rot (DA) & $71.91$ & $74.50$ & $77.94$ & $40.32$ & $ 43.70$ & $46.97$ \\
            + Rot (MT) & $71.63$ & $74.26$ & $78.02$ & $39.22$ & $43.43$ & $46.76$ \\
            + Rot (PT) & $75.86$ & $81.13$ & $84.90$ & $43.08$ & $47.67$ & $52.81$ \\
            + SSP~\cite{yang2020rethinking} & $77.83$ & $82.13$ & - & $43.43$ & $47.11$ & -\\
            + SLA+SD~\cite{lee2020self} & $80.24$ & - & - & $45.53$ & - & - \\
            \hline
            % LDAM-DRW + GridCutRot & 81.2(+5.41\%) & 45.59(+8.44\%) \\
            + LoRot-I& $\underline{81.13}$ & $\underline{83.69}$ & $\underline{86.52}$ & $\underline{45.82}$ & $\underline{49.33}$ & $\textbf{54.69}$\\
            + LoRot-E& $\textbf{81.82}$ & $\textbf{84.41}$ & $\textbf{86.67}$ & $\textbf{46.48}$ & $\textbf{50.05}$ & $\underline{54.66}$\\
            
            % + Rot (DA) & $71.91_{(-6.6\%)}$ & $74.50_{(-8.0\%)}$ & $77.94_{(-8.8\%)}$ & $40.32_{(-4.1\%)}$ & $ 43.70_{(-5.3\%)}$ & $46.97_{(-11.8\%)}$ \\
            % + Rot (MT) & $71.63_{(-7.0\%)}$ & $74.26_{(-8.3\%)}$ & $78.02_{(-8.7\%)}$ & $39.22_{(-6.7\%)}$ & $43.43_{(-5.9\%)}$ & $46.76_{(-12.2\%)}$ \\
            % + Rot (PT) & $75.86_{(-1.5\%)}$ & $81.13_{(+0.2\%)}$ & $84.90_{(-0.7\%)}$ & $43.08_{(+2.5\%)}$ & $47.67_{(+3.3\%)}$ & $52.81_{(-0.8\%)}$ \\
            % + SSP~\cite{yang2020rethinking} & $77.83_{(+1.0\%)}$ & $82.13_{(+1.5\%)}$ & - & $43.43_{(+3.3\%)}$ & $47.11_{(+2.1\%)}$ & -\\
            % + SLA+SD~\cite{lee2020self} & $80.24_{(+4.2\%)}$ & - & - & $45.53_{(+8.3\%)}$ & - & - \\
            % \hline
            % % LDAM-DRW + GridCutRot & 81.2(+5.41\%) & 45.59(+8.44\%) \\
            % + LoRot-I& $\underline{81.13}_{(+5.3\%)}$ & $\underline{83.69}_{(+3.4\%)}$ & $\underline{86.52}_{(+1.2\%)}$ & $\underline{45.82}_{(+9.0\%)}$ & $\underline{49.33}_{(+6.9\%)}$ & $\textbf{54.69}_{(+2.7\%)}$\\
            % + LoRot-E& $\textbf{81.82}_{(+6.2\%)}$ & $\textbf{84.41}_{(+4.3\%)}$ & $\textbf{86.67}_{(+1.4\%)}$ & $\textbf{46.48}_{(+10.6\%)}$ & $\textbf{50.05}_{(+8.5\%)}$ & $\underline{54.66}_{(+2.7\%)}$\\
            \hlineB{2.5}
        \end{tabular}
    }
\end{table*}
\endgroup
\subsubsection{Imbalanced Classification.}
\label{subsection:ImbExperiments}
Following \cite{cao2019learning}, we use CIFAR to design imbalanced scenarios. 
To make imbalanced set, $\upsilon \in (\mu, 1)^{K}$ is multiplied to define the sample numbers for each class as $n_{i} = n_{i}\upsilon_{i}$ where $i$ and $n$ are the class index and the number of the original train set. 
$\mu$ and $K$ denote imbalance ratio and number of classes, respectively.
Then, we measure the accuracy using the original test set. 
% As the baseline, we deploy LDAM-DRW~\cite{cao2019learning} which is one of the SOTA methods in imbalanced classification. 
% We use ResNet32 and follow experimental configurations as done in LDAM-DRW.
As the baseline, we deploy LDAM-DRW~\cite{cao2019learning} and follow experimental configurations from them.
Meanwhile, to compare ours with other self-supervision techniques, 
we also report the results of Rotation~\cite{gidaris2018unsupervised}, SLA+SD~\cite{lee2020self}, and SSP~\cite{yang2020rethinking}. 
To be specific, we apply rotation in the form of DA, MT, and PT as described in Fig.~\ref{figure_CAM}. SSP~\cite{yang2020rethinking} is the method of pre-training the network with self-supervised learning.

In Tab.~\ref{table_IM0.01}, we show LoRot has clear complementary effects and consistently improves the SOTA model by a large gain of up to +4.44\%p (10.56\%) in the highly imbalanced scenario in CIFAR100.
As an analysis, the classifier might not learn the discriminative parts for specific classes only with a few examples in an imbalanced setting since the classifier has a bias towards a small number of samples for such categories.
However, LoRot alleviates this issue since LoRot complements the classifier by discovering sub-discriminative features.
% 
% On the other hand, although rotation has been proved to enhance many aspects of robustness, it does not show improvement against the data imbalance. 
More results with the fully supervised baseline are in the supplementary report.
% Although rotation has been proved to enhance many aspects of robustness, 
% it does not show much improvement against the data imbalance. 
% On the other hand, our methods have clear complementary effects with the LDAM-DRW. Tab.~\ref{table_IM0.01} shows that both LoRot-I and LoRot-E are effective in the imbalanced setting. 
% Particularly, LoRot-E improves the performance based on LDAM-DRW by a large gain of up to 10.56\% in both datasets. More results with the fully supervised baseline is in the Appendix.

\subsubsection{Adversarial Perturbations.}
\label{subsection:AAExperiments}
Substantial efforts were put into improving DNN's robustness~\cite{madry2017towards, athalye2018obfuscated, dong2020benchmarking} to compensate for the vulnerability against adversarial noise~\cite{szegedy2013intriguing}.
For the evaluation, we adopt the PGD training~\cite{madry2017towards} as the baseline following the settings from Rotations~\cite{hendrycks2019using}. 
We conduct experiments on CIFAR10 against $\ell_{\infty}$ perturbations with $\epsilon$ set to 8/255. We adversarially train the network with 10-step adversaries and use 20-step and 100-step adversaries. We set the $\alpha$ to 2/255 for 10, 20-step and 0.3/255 for 100-step as in \cite{madry2017towards, hendrycks2019using}.
% \textbf{Result and Analysis.}
% \begin{wraptable}{hr}{0.5\textwidth}
% 	\centering
% 	{\small
% 		\begin{tabular}{l | c c c}
% 		    \hlineB{2.5}
% 		  %  \multicolumn{5}{|c|}{\textbf{Review of learning needs}} & \multicolumn{4}{c|}{\textbf{Development plan}}\\ \hline
% 		    \multicolumn{1}{c}{} & Clean & 20-step & 100-step \\
% 		  %  \cmidrule(r){2-3} \cmidrule(r){4-5}
% 		    \hlineB{2.5}
%             Baseline & 95.3 & 0.0 & 0.0 \\
%             Adv. Training & 83.4 & 46.5 & 46.5 \\
%             + Rotations~\cite{hendrycks2019using} & 83.5 & 50.3 & 50.4 \\
%             \hline
%             + PatchRot(\textbf{Ours}) & 82.1 & 52.7 & 52.6\\
%             + PatchRot-G(\textbf{Ours}) & 82.6 & 52.8 & 52.8\\
%             \hlineB{2.5}
% 		\end{tabular}
% 	}
% 	\caption{Classification accuracy (\%) against the adversarial attack on CIFAR10. The results show that our model outperforms the baselines in 20-step PGD and 100-step PGD with less degradation of the accuracy for the clean dataset.}
% 	\label{table_AA}
% % \end{table}
% \end{wraptable}
% \endgroup
Tab.~\ref{table_AA} shows the results of LoRot along with the rotation task under the same codebase. Using LoRot led the network to be robust with the increase in PGD attacks by large improvement compared to the baselines. Note that the tradeoff between accuracy and robustness against adversarial noise is very natural~\cite{zhang2019theoretically}.

\begingroup
\setlength{\tabcolsep}{6.0pt} % Default value: 6pt
\renewcommand{\arraystretch}{0.7} % Default value: 1
\begin{table}[t]
    \centering
    {
    	\caption{Classification accuracy (\%) against the adversarial attack on CIFAR10. The results show that our model outperforms the baselines in 20-step PGD and 100-step PGD with less degradation of the accuracy for the clean dataset.}
	\label{table_AA}
		\begin{tabular}{l | c c c}
		    \hlineB{2.5}
		    \multicolumn{1}{l|}{Method} & Clean & 20-step & 100-step \\
		    \hlineB{2.5}
            Baseline & 95.3 & 0.0 & 0.0 \\
            Adv. Training & 83.4 & 46.5 & 46.5 \\
            % + Rotations~\cite{hendrycks2019using} & 83.5 & 50.3 & 50.4 \\
            + Rotations~\cite{hendrycks2019using} & 82.8 & 49.3 & 49.2 \\
            \hline
            + LoRot-I (\textbf{Ours}) & 82.1 & \underline{52.7} & \underline{52.6}\\
            + LoRot-E (\textbf{Ours}) & 82.6 & \textbf{52.8} & \textbf{52.8}\\
            \hlineB{2.5}
		\end{tabular}
	}
\end{table}
\endgroup
% \begin{table}[t]
%     \centering
%     {\small
%         \begin{tabular}{l | c c l}
% 		    \hlineB{2.5}
% 			Method & CIFAR10 & CIFAR100 & ImageNet \\
% 			\hlineB{2.5}
% 			Baseline & 95.01 & 75.07 & 76.32 \\
% 			SLA+SD~\cite{lee2020self} & 94.64 & 76.30 &  $76.17_{(75.17)}$\\
%             \hline
% 			LoRot-I & \textbf{96.16} & \textbf{76.60} & $\underline{77.71}_{(76.32)}$ \\
% 			LoRot-E & \underline{95.96} & \underline{76.36} & $\textbf{77.72}_{(76.32)}$ \\
%             \hlineB{2.5}
% 		\end{tabular}
% 	}
% 	\caption{Top-1 classification accuracy on the CIFAR-10/100 and ImageNet datasets. Numbers in the parenthesis are the baseline accuracy. All experiments are conducted using ResNet50 backbone.}
%     \label{table_Sup_CIFAR}
% \end{table}

\begingroup
\setlength{\tabcolsep}{6.0pt} % Default value: 6pt
\renewcommand{\arraystretch}{0.7} % Default value: 1
\begin{table}[t]
    \centering
    {
    % \caption{ImageNet accuracy (\%). Number in the parenthesis are the baseline accuracy.}
    	\caption{Top-1 and Top-5 Classification accuracy (\%) on ImageNet. Numbers in the parenthesis are the baseline accuracy.}
    \label{table_Sup_CIFAR}
        \begin{tabular}{l | c| c c}
		    \hlineB{2.5}
			Method & Backbone & Top-1 & Top-5 \\
			\hlineB{2.5}
			Baseline & ResNet50 & 76.32 & 92.95 \\
			+ Rot(DA) & ResNet50 & 76.42 & 93.06 \\
			+ Rot(MT) & ResNet50 & 76.68 & 93.10 \\
		    + Rot(SS) & ResNet50 & 76.79 & 93.16 \\
			
			SLA+SD~\cite{lee2020self} & ResNet50 &  $76.17_{(75.17)}$ & - \\
            \hline
			LoRot-I (\textbf{Ours}) & ResNet50 & 77.71 & 93.60 \\
			LoRot-E (\textbf{Ours}) & ResNet50 & \underline{77.72} & \underline{93.65}\\
			\hline
			Cutout~\cite{devries2017improved} & ResNet50 & 77.07 & 93.34 \\
			CutMix~\cite{yun2019cutmix} & ResNet50 & \textbf{78.60} & \textbf{94.08} \\
            \hlineB{2.5}
		\end{tabular}
	}
\end{table}
\endgroup

\subsection{Generalization Capability}
\label{subsection:Classification}
\subsubsection{Image Classification.}
To validate LoRot's benefits in terms of the generalization capability, we evaluate on ImageNet~\cite{deng2009imagenet} and CIFAR datasets.
We compare ours with rotation~\cite{gidaris2018unsupervised} in multiple forms, SLA+SD~\cite{lee2020self}, and patch-based augmentations~\cite{devries2017improved, yun2019cutmix}.
SLA+SD augments the class label by applying rotation and utilizes self-distillation to yield a similar output to ensemble results at inference time.
Tab.~\ref{table_Sup_CIFAR} shows that LoRot clearly achieves the best performance among the methods utilizing rotation. 
Furthermore, we newly spotlight the potential of self-supervision in the perspective of generalization capability in that the gap between LoRot and popular patch-based augmentation, CutMix, is less than 1\% on ImageNet while achieving robustness multifariously.

We further apply our LoRot with data augmentation techniques and contrastive learning. 
Particularly, we test with AutoAugment~\cite{cubuk2019autoaugment}, RandAugment~\cite{cubuk2020randaugment}, and Mixup~\cite{zhang2017mixup} on ImageNet~\cite{deng2009imagenet} and SupCLR~\cite{khosla2020supervised} of contrastive learning on CIFAR datasets. 
% For these experiments, we set our hyperparameter $\lambda$ to 0.1. 
The results in Tab.~\ref{table_ImageNet_Aug} show the consistent trend of the performance gain with three data-augmentation methods without a large number of additional parameters ($+0.12\%$) and extra training time ($+6\%$).
% Note that LoRot can be implemented orthogonally with augmentation techniques 
% To validate the compatibility of LoRot, we applied our self-supervision in the form of multi-task learning while data augmentation can be implemented orthogonally.
% As the results show, without a large number of additional parameters ($+0.12\%$), we observe a steady enhancement in the accuracy. 
Interestingly, we notice that Mixup~\cite{zhang2017mixup} better fits to LoRot-I while Auto- and Rand-Augment are better with LoRot-E.
In the viewpoint of LoRot-E, we speculate that this is because Auto- and Rand-Augment provide the randomness to the grid layout which results in more diverse inputs while Mixup causes a large modification to the image when used with LoRot-E.
Note that LoRot's limitation is that it does not bring surplus benefits to CutMix~\cite{yun2019cutmix}~($\pm 0\%$) since LoRot and patch-based augmentations may modify overlapped region and interrupt each other.
% However, there is no significant improvement since our self-supervision is also based on regional modification.
% Note that Mixup~\cite{zhang2017mixup} and CutMix~\cite{yun2019cutmix} modify much part of the image but achieve outstanding accuracy since they smooth the label information as much as they change the image.
% Note that Patch-Rot is similar to regional dropout methods so that the performances of regional dropout methods were not improved.
%CutOut~\cite{devries2017improved} drops out randomly sampled region in the image and Mixup~\cite{zhang2017mixup} interpolates pair of images to extend training distribution. Autoaugment~\cite{cubuk2019autoaugment} designs augmentation policy with reinforcement leraning and its performance is from the original paper. To validate the compatibility of Patch-Rot, we applied our methods as auxiliary self-supervision to above augmentation techniques. As the results show, without much increase in the number of parameters($+0.12\%$), ours can improve classification accuracy of these methods. Note that Patch-Rot is similar to regional dropout so that regional dropout methods are not compatible with our methods.
% \textbf{Contrastive Learning}
% Contrastive Learning is one of the spotlighted training schemes these days and has achieved promising results for both unsupervised~\cite{chen2020simple, caron2020unsupervised} and supervised learning~\cite{khosla2020supervised}. 

Contrastive Learning has achieved promising results for both unsupervised~\cite{chen2020simple, caron2020unsupervised} and supervised learning~\cite{khosla2020supervised}. 
% As it is proved that self-supervision techniques can be enhanced by combining relation-based and transform-based methods in PIRL~\cite{misra2020self}, we also examine it in supervised domain. 
As it is shown that relation-based and transform-based methods are complementary in PIRL~\cite{misra2020self}, we also examine it in terms of supervised domain. 
% contrastive learning and transform-based pretext tasks can be further improved by adopting one another.
% Not only multi-image based self-supervised learning has achieved comparable performances to fully supervised setting, but also show promising results when modified for supervised learning.
% Specifically, we attach LoRot to SupCLR~\cite{khosla2020supervised}.
We report the performance of SupCLR~\cite{khosla2020supervised} both from its paper and our reproduced version in Tab.~\ref{table_Sup_CIFAR_complementary}. 
We first applied rotation~\cite{gidaris2018unsupervised} with two different strategies: MT and PT.
However, applying rotation with MT provoked the decline in the performance as mentioned in contrastive learning~\cite{chen2020learning} that rotation as augmentation degrades the discriminative performance. 
As is, using rotation only for self-supervised loss (PT) was not very efficient either, in that it requires twice more computational cost to yield insignificant increase.
On the contrary, LoRot benefits additive effects to contrastive learning by enriching the representation vectors that are to be pushed or pulled between other samples.
% See Tab.~\ref{table_Sup_CIFAR_complementary} for the result.
% \paragraph{Setup} To integrate Patch-Rot with the supervised contrastive learning, we exploited contrastive loss to conduct multi-task learning in the first phase without using classifier. 
% Then, we finetuned single classifier as SupCLR~\cite{khosla2020supervised} did.
% For the CIFAR10/100 datasets, we followed the settings from SupCLR except that we set the batch size to 512 for our experiments.

\begingroup
\setlength{\tabcolsep}{6pt} % Default value: 6pt
\renewcommand{\arraystretch}{0.8} % Default value: 1
\begin{table}[t]
    \centering
    {
    	\caption{Additive benefits of LoRot with augmentation methods on ImageNet classification (\%). LoRot shows a consistent trend of performance gains.}
	\label{table_ImageNet_Aug}
	    \begin{tabular}{l | c | c c | c c|  c c}
		    \hlineB{2.5}
		    \multirow{2}{*}{Method} & \multirow{2}{*}{Backbone} & \multicolumn{2}{c|}{-} & \multicolumn{2}{c|}{+LoRot-I}& \multicolumn{2}{c}{+LoRot-E} \\ %\hline
			& & Top-1 & Top-5& Top-1 & Top-5& Top-1 & Top-5   \\
			\hlineB{2.5}
			Mixup~\cite{zhang2017mixup} & ResNet50 &77.58 & 93.60 &\textbf{78.36} & \textbf{94.15} &\underline{78.18} & \underline{94.05}\\
			\hline
			AutoAug~\cite{cubuk2019autoaugment} & ResNet50 &77.60 & 93.80 &\underline{78.09} & \underline{93.76} &\textbf{78.22} & \textbf{93.86} \\
			\hline
			RandAug~\cite{cubuk2020randaugment} & ResNet50 &77.52 & 93.47 &\underline{78.12} & \underline{93.84} &\textbf{78.24} & \textbf{93.95} \\
% 			\hline
            \hlineB{2.5}
		\end{tabular}
	}
\end{table}
\endgroup
\begingroup
\setlength{\tabcolsep}{7pt} % Default value: 6pt
\renewcommand{\arraystretch}{0.7} % Default value: 1
\begin{table}[t]
    \centering
    {
    	\caption{Additive benefits of LoRot with contrastive learning on CIFAR-10/100 classification and OOD detection. `$\dagger$' indicates the number taken from the paper~\cite{khosla2020supervised} using batch size of 1024. The rest of the results were reproduced with batch size of 512 due to lack of GPU memory. OOD scores are measured with the trained model on CIFAR-10 and averaged over the datasets in Table.~\ref{table_OOD}. All results are averaged on three trials.}
	\label{table_Sup_CIFAR_complementary}
		\begin{tabular}{l | c c c c c}
		    \hlineB{2.5}
			Method & CIFAR10 & CIFAR100 & OOD  \\
			\hlineB{2.5}
			\textcolor{gray}{SupCLR~\cite{khosla2020supervised}$\dagger$} &\textcolor{gray}{96.0} & \textcolor{gray}{76.5} & \textcolor{gray}{N/A}  \\
			SupCLR~\cite{khosla2020supervised} & 95.75 & 76.52 &  96.98\\
			+ Rotation (MT) & 94.24 & 71.80 & 96.28  \\
			+ Rotation (PT) & 96.07 & 76.73 & 96.90 \\
			\hline
			+ LoRot-I & \textbf{96.79} & \textbf{78.78} & \textbf{97.95} \\
			+ LoRot-E & \underline{96.73} & \underline{78.77} & \underline{97.92}\\
% 			SupCLR+GridCutRot & ResNet50 &  & 77.61 & 100+5M \\
            \hlineB{2.5}
		\end{tabular}
	}
\end{table}
\endgroup
% \textbf{Results and Analysis.}

% To measure the OOD score, we used the trained model on CIFAR-10 dataset. 
% Detailed results are in the Supplementary material.

% \begin{table}[!t]
%     \centering
%     \small
%     \begin{minipage}[t]{0.36\linewidth}%\centering
%     \setlength{\tabcolsep}{6pt} % Default value: 6pt
	
%     \end{minipage}\hfill%
%     \begin{minipage}[t]{0.62\linewidth}\centering
%     \setlength{\tabcolsep}{6pt} % Default value: 6pt

% \end{table}

\subsubsection{Localization and Transfer Learning}
\label{subsection:highlevelunderstanding}
 are important criteria to evaluate the model's localization capability. 
For these experiments, we used our pretrained model yielded from Tab.~\ref{table_Sup_CIFAR}.
Briefly, for weakly supervised object localization, the model needs to localize the object when only given with class labels.
Thus, the model is required to not only find the class-descriptive clues but also understand the image.
Tab.~\ref{table_wsol} demonstrates that LoRot better guides the model to focus on salient regions. Particularly, we observe that LoRot-E, explicitly having the localization task, leads to better localization capability. For evaluation, we use CAM~\cite{zhou2016learning} following ACOL~\cite{zhang2018adversarial} and Co-mixup~\cite{kim2021co}. As ACOL searched for threshold for CAM results between 0.5 to 0.9, we report all these results. 
% For the baseline, we use publicly available models\footnote{https://github.com/clovaai/CutMix-PyTorch} that are trained with the same protocol.
\begin{table}[!t]
    \centering
    \small
    \begin{minipage}[t!]{0.54\linewidth}%\centering
    \setlength{\tabcolsep}{3pt} % Default value: 6pt
    \renewcommand{\arraystretch}{0.7} % Default value: 1
        	\centering
        	{\small
            \caption{Weakly Supervised Object Localization accuracy (\%) on ImageNet.}
        	\label{table_wsol}
            \begin{tabular}{l | c c c c c}
    		    \hlineB{2.5}
                Threshold & 0.5 & 0.6 & 0.7 & 0.8 & 0.9 \\
    			\hlineB{2.5}
    			Baseline & 46.72 & 31.55 & 14.49 & 4.22 & 1.91 \\
    			CutMix & 47.39 & 30.24 & 13.86 & 4.57 & \underline{2.03} \\
    			LoRot-I & \underline{49.73} & \underline{35.49} & \underline{17.21} & \underline{5.08} & \underline{2.03} \\
    			LoRot-E & \textbf{50.24} & \textbf{36.07} & \textbf{17.81} & \textbf{5.49} & \textbf{2.12} \\
                \hlineB{2.5}
    		\end{tabular}
        	}
    \end{minipage}\hfill%
    \begin{minipage}[t!]{0.44\linewidth}
    \setlength{\tabcolsep}{4pt} % Default value: 6pt
    \renewcommand{\arraystretch}{0.7} % Default value: 1
    	\centering
    	{\small
        \caption{AP (\%) of object detection and instance segmentation models initialized with each pretrained method.}
    	\label{table_transferlearning}
		\begin{tabular}{l | c c}
		    \hlineB{2.5}
		    Pretrained & RetinaNet & SOLOv2 \\\hline
			Baseline & 33.8 & 33.7 \\
			LoRot-I & \textbf{35.3} & \textbf{34.5} \\
			LoRot-E & \underline{35.2} & \underline{34.4} \\
            \hlineB{2.5}
		\end{tabular}
    	}
    \end{minipage}%\hfill
\end{table}

% \textbf{Transfer learning with pretrained models}
Object detection and instance segmentation are another tasks that require precise localization capability of the model. 
Indeed, backbones are commonly initialized with ImageNet pretrained weights to deal with the lack of labeled train data. 
Thus, we examine whether pretrained models trained with LoRot yield any benefits. 
For evaluation, we employ Retinanet~\cite{lin2017focal} and SOLOv2~\cite{wang2020solov2} for each task and use COCO 2017 dataset~\cite{lin2014microsoft} for experiments. Tab.~\ref{table_transferlearning} shows our findings: pretrained models with LoRot consistently outperform standard models. 

% \textbf{implementation details}
% For instance segmentation, we use real-time SOLOv2 model where the number of convolution layers in the prediction head is reduced to two and the input shorter side is 448. We train the model with the 3x schedule as reported in their paper. For retinanet, we train the model with the 1x schedule and MS COCO 2017 dataset~\cite{lin2014microsoft} is used for both experiments.

% instance segmentation - MS COCO 2017
% SOLOv2 light 448 R50 3x - real time SOLO v2 (speed priority, Real-time setting We design two light-weight models for different purposes. 1) Speed priority,
% the number of convolution layers in the prediction head is reduced to two and the input shorter side is
% 448.)

% Object detection - retinanet - MS COCO 2017 validation set
% R50 - retinanet R50 FPN1x.yaml
% \begingroup
% \setlength{\tabcolsep}{12pt} % Default value: 6pt
% \renewcommand{\arraystretch}{1.0} % Default value: 1

%% file: _5_further_analysis.tex
\subsection{Further Study}
\label{subsection:furtherstudy}
\begin{figure*}[t!]
    \centering
    % \vspace{-0.3cm}
    \includegraphics[scale=0.4]{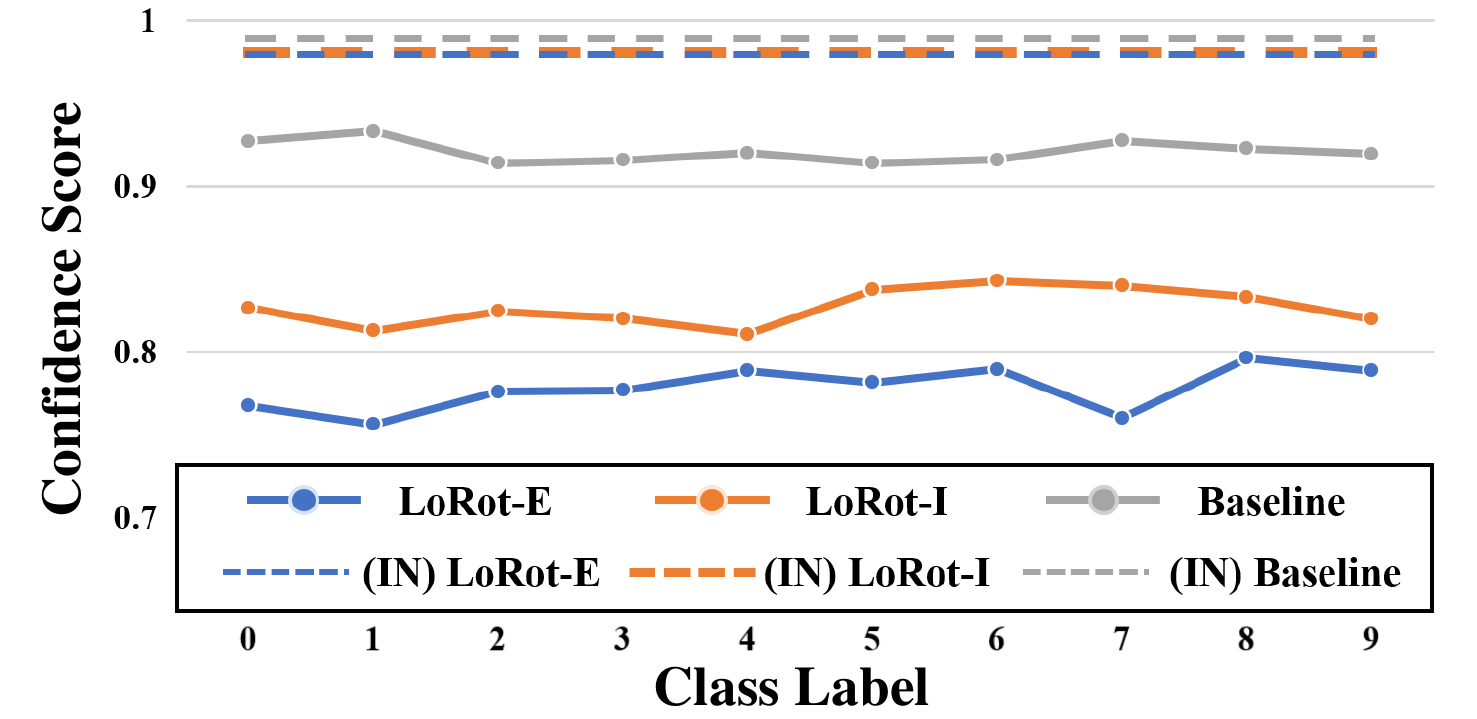}
    % \vspace{-0.33cm}
    \caption{Average confidence scores for in- and out-of-distribution data (CIFAR10 and SVHN) of the baseline and LoRot.
    Dotted lines are the averaged confidence scores of in-distribution (IN-) dataset for each method. Solid lines represent the confidence scores (y-axis) for each class in out-distribution dataset (x-axis). These results demonstrate that LoRot improves the capability of the models to detect unknown samples.}
    %the model's unknown detecting capability across novel classes.}
    %Class-by-class performance analysis with confidence scores from the baseline and LoRot.
    \label{figure_OOD_conf}
    % \vspace{-0.45cm}
\end{figure*}

To understand why LoRot is effective in enhancing robustness, we conducted an in-depth analysis of OOD detection shown in Tab.~\ref{table_OOD}.
% To be specific, we measured the confidence gaps between in- and out-distribution test sets.
In Fig.~\ref{figure_OOD_conf}, we compare the class-wise average confidence scores for OOD (SVHN) dataset between the baseline and the LoRot.
We observe a clear tendency that both the LoRot-I and LoRot-E effectively lower all the confidence scores for OOD classes while retaining high confidence for in-distribution dataset.
Therefore, LoRot can achieve higher AUROC scores.
% This is why a larger AUROC value has been obtained from the ROC curve for the proposed method. 

\begin{figure*}[t]
    \centering
    % \vspace{0.05cm}
    \includegraphics[scale=0.4]{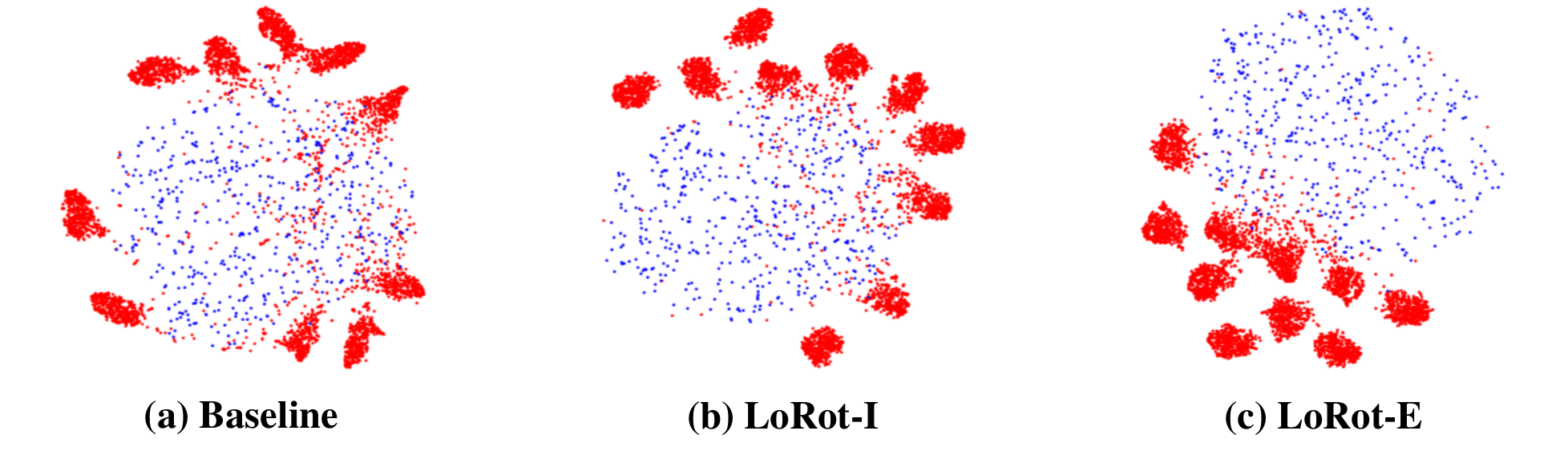}
    % \vspace{-0.3cm}
    \caption{From left to right, t-SNE visualization for the baseline, LoRot-I, and LoRot-E, respectively. We plot the feature distributions of in-distribution instances (Red) and out-distribution instances (Blue). Unlike the baseline where many red dots are scattered with the blue dots, it is evident that clusters appearing under LoRot is more compact.}
    \label{figure_OOD_TSNE}
    % \vspace{-0.55cm}
\end{figure*}

Furthermore, we visualized the final embedding space with t-SNE to explore underneath reason for why a better separation has been achieved by the proposed method.
In Fig.~\ref{figure_OOD_TSNE}, ten red clusters and blue dots can be found for three methods.
Red clusters represent classes in in-distribution dataset, CIFAR10, and 
% For all cases, ten red clusters can be found in the visualization plane in Fig.~\ref{figure_OOD_TSNE}; each of these represents each class in in-distribution dataset, CIFAR10.
blue dots are embeddings of OOD dataset, SVHN.
Yet, we can observe that the red and blue dots are significantly mixed in the baseline's embedding space.
% Yet, we can observe considerable overlap between red and blue dots in the baseline's feature space. 
Meanwhile, two colors overlie less on top of the other in LoRot's feature space and tighter boundaries are formed for red clusters.
As discussed, this observation is because the model obtains rich features through learning LoRot which enables the model to understand the input even when the most discriminative hint for each class is not available.
% In other words, the classifier is not vulnerable to predicting OOD samples as one of the known classes even with discriminatory features of some classes.
In other words, the model is less vulnerable to mispredicting OOD samples with learned features of some classes because it considers a broader spectrum of class-descriptive features.
% In other words, the model is less vulnerable to predicting OOD sample with learned features of some classes as known because it also considers less discriminative features.
% Consistent results on other OOD datasets are available in the supplementary.

%% file: _6_conclusion.tex
\section{Conclusion}
Although self-supervision has been proved to be powerful in supervised domain, its potential is still an untapped question since existing works are designed for unsupervised condition.
Thus, we presented three desirable properties of self-supervision to be tailored for supervised learning: enriching representations, maintaining data distribution, and high applicability.
To comply with them, we introduced LoRot, a self-supervised localization task that assists supervised learning to further improve robustness and generalization capability.
Our extensive experiments demonstrated the merits of LoRot as well as the complementary benefits to prior arts. 
Furthermore, as we revisited the potential of self-supervision in a simple applicable way in supervised settings, we believe this line is worth further study to be a standard technique in supervised learning.

\vspace{5pt}
\noindent\textbf{Acknowledgements.} 
This work was supported in part by MSIT/IITP (No. 2022-0-00680, 2020-0-00973, 2020-0-01821, and 2019-0-00421), MCST/KOCCA (No. R2020070002), and MSIT\&KNPA/KIPoT (Police Lab 2.0, No. 210121M06).

%% file: Supplementary/abstract.tex
% \begin{outline}
\vspace{-0.2cm}
As elaborated in the main paper, our proposed LoRot is a self-supervision task tailored for supervised learning.
Our motivation is that self-supervisions previously adopted in supervised learning were originally designed for unsupervised representation learning, thus significant extra computational costs for training were required to achieve insignificant gains.
To maximize the benefits for supervised learning, we first introduced three desirable properties of self-supervision and how can pretext tasks can satisfy these conditions by proposing LoRot.
To learn rich features, LoRot discovers subdiscriminative features within the part of the image that are not usually considered by current supervised models.
Also, LoRot only rotates a part of the image which does not make much changes within the image.
Lastly, LoRot is utilized in the form of multi-task learning to have high efficiency.
%that are efficient and effective.
% First, the pretext tasks need to guide the model learn rich features. 
% Second, the data transformations for pretext tasks should not significantly alter the training distribution. 
% Finally, the self supervised tasks are preferred to be light and generic for high applicability to prior arts. 
% In this context, we proposed LoRot, the tailored self supervision for supervised learning. 
% Briefly, LoRot only rotates a part of the image and make quizzes within the parts.
% In this way, LoRot satisfies above conditions by not incurring much changes in the image and discovering subdiscriminative features that are not considered by the current supervised models.
In this supplementary report, we provide ablation and further studies of LoRot, as following outline.
%For the contents, We give the outline as below:

\Skip{
            As elaborated in the main paper, our proposed LoRot is a tailored self-supervision for supervised learning.
            Our motivation was that self-supervisions in supervised learning was originally designed for unsupervised representation learning so that significant extra computational costs were required to achieve insignificant gains.
            To maximize the benefits for supervised learning, we first introduced three desirable properties of self-supervision and proposed LoRot that satisfies all conditions.
            To learn rich features, LoRot discovers subdiscriminative features within the part of the image that are not usually considered by current supervised models.
            Also, LoRot only rotates a part of the image which does not make much changes within the image.
            Lastly, LoRot is utilized in the form of multi-task learning that are efficient and effective.
            % First, the pretext tasks need to guide the model learn rich features. 
            % Second, the data transformations for pretext tasks should not significantly alter the training distribution. 
            % Finally, the self supervised tasks are preferred to be light and generic for high applicability to prior arts. 
            % In this context, we proposed LoRot, the tailored self supervision for supervised learning. 
            % Briefly, LoRot only rotates a part of the image and make quizzes within the parts.
            % In this way, LoRot satisfies above conditions by not incurring much changes in the image and discovering subdiscriminative features that are not considered by the current supervised models.
            Hereby, we provide additional experiments, ablation studies, and detailed analysis of LoRot.
            For the contents, We give the outline as below:
}
    \vspace{5pt}
    
    \textbf{Part \ref{section:Further Analysis}: Ablation study on the hyperparameter $\lambda$}
    
    \textbf{Part \ref{section:patch size}: Ablation study / analysis on the patch sizes for LoRot-I}
    
    \textbf{Part \ref{section:spatialpooling}: Ablation study on spatial pooling methods for LoRot-E}
    
    \textbf{Part \ref{OODwithCSI}: Detailed results on OOD detection with SupCLR~\cite{khosla2020supervised}}
    
    \textbf{Part \ref{imbalanced_baseline}: Results on imbalanced classification with the baseline}
    
    \textbf{Part \ref{section:OODanalysis}: Further study of LoRot on other datasets in OOD detection}
    
    % \textbf{Section \ref{subsection:OSRExperiments} : Additional Experiments on Open set recognition} upon Mixmatch
    
    % \textbf{Section \ref{supp:additional} : Additional Experiments semi-supervised learning}
    
    \textbf{Part \ref{supp:implementationdetails}: Implementation details}
    
    \vspace{5pt}
    
    \noindent Throughout this supplementary report, bolds and underlines in tables indicate the best and the second best scores, respectively.
    Also note that colored references, e.g. \textcolor{blue}{Tab., Fig.} denotes table and figure in the main paper. 
    
% \end{outline}

%% file: Supp_lambdaExp.tex
%\section{Effect of $\lambda$ on the Performance}
\section{Effect of $\lambda$ in Objective Function}
\label{section:Further Analysis}
In Tab.~\ref{table_lambda}, we report the performances with varying $\lambda$ which controls the loss ratio between the primary objective and our self-supervision task. 
We found that $\lambda$ does not lead significant performance variations, while $\lambda = 0.2$ usually provides higher scores than the second-best among tested methods in all the tasks. 
For the classification tasks on CIFAR datasets, we ran experiments under the same setting with \textcolor{blue}{Tab. 1} in our main paper.

% 0.2, 0.3 일때는 sla , ood 방법들잡는다 기술.
\begingroup
\setlength{\tabcolsep}{10pt} % Default value: 6pt
\renewcommand{\arraystretch}{1.2} % Default value: 1
\begin{table}[h]
% 	\vspace{0.2cm}
	\centering
		\caption{Classification accuracies (\%) on CIFAR datasets and AUROC (\%) scores of OOD detection with varying $\lambda$. The reported classification and OOD results are averaged over 3 and 5 runs on all the datasets, respectively. Additionally, we report the performances of a fully supervised baseline for comparison. Results demonstrate that LoRot is not very sensitive to hyperparameter that it outperforms the baseline by large margins regardless of the value of $\lambda$.}
	\label{table_lambda}
	{\small
		\begin{tabular}{l | c | c c c}
		    \hlineB{2.5}
		    & $\lambda$ & CIFAR10 & CIFAR100 & OOD  \\
		    \hlineB{2.5}
		    Baseline & - & 95.01 & 75.07 &  86.07 \\
		    \hline
            \multirow{5}{*}{LoRot-I} & 0.1 & 95.92 & 76.49 & 94.55 \\
            & 0.2 & 96.16 & 76.60 & 95.20 \\
            & 0.3 & 95.72 & 76.57 & 94.88 \\
            & 0.4 & 95.92 & 75.97 & 94.98 \\
            & 0.5 & 95.84 & 75.95 & 94.90 \\
            \hline
            \multirow{5}{*}{LoRot-E} & 0.1 & 95.77 & 75.9 & 94.83 \\
            & 0.2 & 95.96 & 76.36 & 94.83 \\
            & 0.3 & 95.75 & 76.4 & 94.63 \\
            & 0.4 & 95.76 & 76.13 & 94.56\\
            & 0.5 & 95.73 & 76.13 & 94.55 \\
            \hline

            \hlineB{2.5}
		\end{tabular}
	}
% 	\vspace{0.2cm}

% 	\vspace{0.2cm}
\end{table}
\endgroup

\Skip{
        $\lambda$ is the hyperparameter to control the loss ratio between the primary objective and the pretext task. In the paper, we simply set the $\lambda$ to either 0.1 or 0.2 for the consistent experimental settings.
        However, we found that $\lambda$ does not make noticeable changes in the performances of LoRot-I.
        In Tab.~\ref{table_lambda}, the model consistently shows improvements over the values of $\lambda$ as we test our model with various $\lambda$ in both OOD detection and image classification.
        In fact, the results when the $\lambda$ is set to 0.2 and 0.3 are better than the second-best methods in each task in the paper.
        % 0.2, 0.3 일때는 sla , ood 방법들잡는다 기술.
        \begingroup
        \setlength{\tabcolsep}{10pt} % Default value: 6pt
        \renewcommand{\arraystretch}{1.2} % Default value: 1
        \begin{table}[h]
        % 	\vspace{0.2cm}
        	\centering
        	{\small
        		\begin{tabular}{l | c | c c c}
        		    \hlineB{2.5}
        		    & $\lambda$ & CIFAR10 & CIFAR100 & OOD  \\
        		    \hlineB{2.5}
                    \multirow{5}{*}{LoRot-I} & 0.1 & 95.92 & 76.49 & 94.55 \\
                    & 0.2 & 96.16 & 76.60 & 95.20 \\
                    & 0.3 & 95.72 & 76.57 & 94.88 \\
                    & 0.4 & 95.92 & 75.97 & 94.98 \\
                    & 0.5 & 95.84 & 75.95 & 94.90 \\
                    \hline
                    \multirow{5}{*}{LoRot-E} & 0.1 & 95.77 & 75.9 & 94.93 \\
                    & 0.2 & 95.96 & 76.36 & 94.83 \\
                    & 0.3 & 95.75 & 76.4 & 94.63 \\
                    & 0.4 & 95.76 & 76.13 & 94.56\\
                    & 0.5 & 95.73 & 76.13 & 94.55 \\
                    \hline

                    \hlineB{2.5}
        		\end{tabular}
        	}
        	\vspace{0.2cm}
        	\caption{Classification accuracy (\%) on CIFAR datasets and AUROC (\%) score for OOD Detection w.r.t $\lambda$. All classification results are averaged over 3 runs and OOD experiments are averaged over 5 runs on all datasets. }
        	\label{table_lambda}
        	\vspace{-0.5cm}
        \end{table}
        \endgroup
}

%% file: Supp_patchsizeExp.tex
% \section{Ablation Study}
\begingroup
\setlength{\tabcolsep}{7pt} % Default value: 6pt
\renewcommand{\arraystretch}{1.2} % Default value: 1
\begin{table*}[ht!]
% 	\vspace{0.2cm}
	\centering
		\caption{Classification accuracies (\%) on CIFAR datasets and AUROC scores (\%) for OOD detection with various patch configurations. When the Min and Max patch sizes differs, we randomly sample the patch size within the range, where W denotes the width of the image. For OOD detection, we follow the standard setting used in the paper and report average AUROC scores. Note that, the top row in the table is the reported results of LoRot-I in the paper. All the classification results are averaged over 3 trials, and OOD experiments are averaged over 5 trials.}
	\label{table_patchsize}
	{\small
		\begin{tabular}{l | c | c | c c c}
		    \hlineB{2.5}
		  %  \multicolumn{5}{|c|}{\textbf{Review of learning needs}} & \multicolumn{4}{c|}{\textbf{Development plan}}\\ 
		    \multirow{2}{*}{} & \multicolumn{2}{c|}{Patch Size} & \multicolumn{3}{c}{Settings} \\
		    \cline{2-6}
		    & Min & Max & CIFAR10 & CIFAR100 & OOD  \\
		    
		    \hlineB{2.5}
		    \multirow{3}{*}{Random Size} & 2 & W / 2 & 96.16 & 76.60 & 94.55 \\ 
            \cline{2-6}
            & 2 & W / 4 & 95.91 & 77.01 & 94.51 \\
            & W / 4 & W / 2 & 95.95 & 75.78 & 95.38 \\
            \hline
            \multirow{3}{*}{Fixed Size} & 2 & 2 & 94.93 & 76.25 & 93.64 \\
            & W / 4 & W / 4 & 95.68 & 76.39 & 94.16 \\
            & W / 2 & W / 2 & 94.45 & 74.98 & 95.06  \\
            % \multirow{3}{*}{Random Size} & 2 & W / 2 & 76.6 & \textbf{96.16} & 94.55 \\ 
            % \cline{2-6}
            % & 2 & W / 4 & \textbf{77.01} & 95.91 & 94.51 \\
            % & W / 4 & W / 2 & 75.78 & 95.95 & \textbf{95.38} \\
            % \hline
            % \multirow{3}{*}{Fixed Size} & 2 & 2 & 76.25 & 94.93 & 93.64 \\
            % & W / 4 & W / 4 & 76.39 & 95.68 & 94.16 \\
            % & W / 2 & W / 2 & 74.98 & 94.45 & 95.06  \\
            \hline

            \hlineB{2.5}
		\end{tabular}
	}
% 	\vspace{0.1cm}

% 	\vspace{-0.5cm}
\end{table*}
% \endgroup
\section{Effect of the Patch Sizes in LoRot-I}
\label{section:patch size}
\begingroup
\setlength{\tabcolsep}{7pt} % Default value: 6pt
\renewcommand{\arraystretch}{1.2} % Default value: 1
\begin{table*}[ht!]
% 	\vspace{0.2cm}
	\centering
		\caption{Comparison of three different spatial feature pooling methods for LoRot-E. Reported results are the classification accuracies (\%) on CIFAR datasets and AUROC scores (\%) for OOD. Note that, $w_{f}$ and ${h}_{f}$ indicate the width and height of the feature map, respectively.}
	\label{table_GAP}
	{\small
		\begin{tabular}{l c | c c c}
		    \hlineB{2.5}
		  %  \multicolumn{5}{|c|}{\textbf{Review of learning needs}} & \multicolumn{4}{c|}{\textbf{Development plan}}\\ \hline
		    Spatial Pooling & Spatial Dim & CIFAR10 & CIFAR100 & OOD  \\
		  %  \cmidrule(r){2-3} \cmidrule(r){4-5}
		    \hlineB{2.5}
            Dense & $w_{f} \times h_{f}$  & 95.76 & 74.5 & 94.20\\ 
            Reduced Dense & $2 \times 2$ & 95.79 & 76.05 & 94.70 \\
            GAP & $1 \times 1$ & \textbf{95.96} & \textbf{76.36} & \textbf{94.90} \\
            \hline

            \hlineB{2.5}
		\end{tabular}
	}
% 	\vspace{0.2cm}

	\vspace{-0.2cm}
\end{table*}
\endgroup

% \endgroup
To investigate the performance variations of our LoRot-I with respect to the patch sizes, we perform experiments with two configurations, the fixed-sized and random-sized patches, and the results are reported in Tab.~\ref{table_patchsize}. Overall, the random-sized patches outperform the fixed-sized ones.
%, as the PatchRot with fixed-sized patches is similar to PatchRot-G without the locational query
% One interesting finding from these experiments is that the patch sizes are related to the trade-off between the robustness and accuracy of the model. 
One interesting finding from Tab.~\ref{table_patchsize} is that there is a trade-off between the robustness and accuracy of the model depending on the size of the patch.
Specifically, the smaller patches lead high accuracy but lower robustness, and bigger ones bring the opposite tendency. 
We think that these are mainly because the bigger patches are highly likely to produce quizzes with regions containing objects that spread out the model's attention, while the smaller patches are more like to work as data augmentation. 
Indeed, spreading the model's attention is more advantageous for the model's robustness since it forces the model to consider sub-discriminative features.
Among various configurations, we choose one that yields a good balance between the accuracy and robustness in the paper.

\Skip
{

        \section{Effect of the Patch Size in PatchRot}
        With the proposed PatchRot, we found out that there is a correlation between the patch size and the trade-off between generalization capability and robustness in our method.
        To be specific, we conducted experiments on two conditions: the fixed-size patch and the random-size patch to test how the patch size affects the model training.
        Tab.~\ref{table_patchsize} shows the results over different sizes of the patch.
        First, we observed that the random-size patch is better than the fixed-size patch. 
        PatchRot with the fixed-size patch is similar to PatchRot-G but without a pretext task for location.
        Then we also discovered that relatively a smaller patch size improves the classification accuracy since it provides more data augmentation.
        In contrast, as the overall patch size gets bigger, PatchRot shows improvements mainly on robustness because they are more likely to generate meaningful quizzes that contain parts of the object.
        % This forces the model to look at more details of the object.
        Among these variants in patch size, we reported PatchRot that yields consistently good results over many tasks.
        
        \begingroup
        \setlength{\tabcolsep}{5pt} % Default value: 6pt
        \renewcommand{\arraystretch}{1.2} % Default value: 1
        \begin{table}[h]
        % 	\vspace{0.2cm}
        	\centering
        	{\small
        		\begin{tabular}{l | c | c | c c c}
        		    \hlineB{2.5}
        		  %  \multicolumn{5}{|c|}{\textbf{Review of learning needs}} & \multicolumn{4}{c|}{\textbf{Development plan}}\\ 
        		    \multirow{2}{*}{} & \multicolumn{2}{c|}{Patch Size} & \multicolumn{3}{c}{Settings} \\
        		    \cline{2-6}
        		    & Min & Max & CIFAR10 & CIFAR100 & OOD  \\
        		    
        		    \hlineB{2.5}
        		    \multirow{3}{*}{Random Size} & 2 & W / 2 & \textbf{96.16} & 76.60 & 94.55 \\ 
                    \cline{2-6}
                    & 2 & W / 4 & 95.91 & \textbf{77.01} & 94.51 \\
                    & W / 4 & W / 2 & 95.95 & 75.78 & \textbf{95.38} \\
                    \hline
                    \multirow{3}{*}{Fixed Size} & 2 & 2 & 94.93 & 76.25 & 93.64 \\
                    & W / 4 & W / 4 & 95.68 & 76.39 & 94.16 \\
                    & W / 2 & W / 2 & 94.45 & 74.98 & 95.06  \\
                    % \multirow{3}{*}{Random Size} & 2 & W / 2 & 76.6 & \textbf{96.16} & 94.55 \\ 
                    % \cline{2-6}
                    % & 2 & W / 4 & \textbf{77.01} & 95.91 & 94.51 \\
                    % & W / 4 & W / 2 & 75.78 & 95.95 & \textbf{95.38} \\
                    % \hline
                    % \multirow{3}{*}{Fixed Size} & 2 & 2 & 76.25 & 94.93 & 93.64 \\
                    % & W / 4 & W / 4 & 76.39 & 95.68 & 94.16 \\
                    % & W / 2 & W / 2 & 74.98 & 94.45 & 95.06  \\
                    \hline

                    \hlineB{2.5}
        		\end{tabular}
        	}
        	\vspace{0.1cm}
        	\caption{Classification accuracy (\%) on CIFAR dataset and AUROC score (\%) for OOD Detection w.r.t various patch sizes. When the Min and Max Patch size differs, we randomly sample the patch size. W denotes the width of the image. For OOD Detection, we follow standard setting in the paper and report average AUROC score. Top row in the table is the reported PatchRot. All classificatio n results are averaged on 3 trials and OOD experiments are averaged on 5 trials.}
        	\label{table_patchsize}
        	\vspace{-0.5cm}
        \end{table}
        \endgroup

        % \newpage
        
        \section{Effect of the Grid Layout in PatchRot-G}
        For PatchRot with a sampling method in a grid, we tested our model to see how the size of the grid and patch affect the model performance. 
        For this experiment, we define a 4x4 grid and conduct an experiment with different patch sizes.
        % All experiments are conducted using multi-task learning strategy.
        As illustrated in the paper, we can see that rotating large portion (3x3 patch in a 4x4 grid) of the image often degrades the performance as they are likely to modify data distribution. 
        On the other hand, defining too small fractions as a target is not very good neither since they have high probability of generating uninformative patches which disturbs the primary objective. 
        In Tab.~\ref{table_gridsize}, 2x2 patch in a 4x4 grid is similar to PatchRot-G in the size ratio of the patch is same.
        However, they are different in that the 2x2 patch in a 4x4 grid can be positioned at 9 locations while the patch from PatchRot-G can only be placed at each quadrant.
        This may the cause for the difference in the performances since a 2x2 patch in a 4x4 grid is more likely to rotate the object when the patch is at the center.
        
        \begingroup
        \setlength{\tabcolsep}{5pt} % Default value: 6pt
        \renewcommand{\arraystretch}{1.2} % Default value: 1
        \begin{table}[h]
        	\centering
        	{\small
        		\begin{tabular}{c c | c c c}
        		    \hlineB{2.5}
        		    Grid Layout & Rotating Cell Size & CIFAR10 & CIFAR100 & OOD \\
        		    \hline
        		    2x2 & 1x1 & \textbf{95.96} & \textbf{76.36} & \textbf{94.93} \\
        		    \hline
        		    4x4 & 1x1 & 95.11 & 76.12 & 93.49 \\
        		    4x4 & 2x2 & 95.88 & 74.71 & 94.57 \\
        		    4x4 & 3x3 & 94.96 & 72.26 & 93.46 \\
        		    \hline
                    \hlineB{2.5}
        		\end{tabular}
        	}
        	\vspace{0.1cm}
        	\caption{Classification Accuracy (\%) on CIFAR dataset and AUROC score (\%) for OOD Detection with different grid sizes. Grid indicates how the single image is divided and the patch size is the size of the target cell which we actually apply the rotation. Note that the top row is PatchRot-G. Reported results are averaged on 3, 5 trials, respectively depending on the task.}
        	\label{table_gridsize}
        	
        \end{table}
        \endgroup
        
}

%% file: Supp_GAP.tex
% \begin{table*}[t!]
% % 	\vspace{0.2cm}
% 	\centering
% 	{\small
% 		\begin{tabular}{l c | c c c}
% 		    \hlineB{2.5}
% 		  %  \multicolumn{5}{|c|}{\textbf{Review of learning needs}} & \multicolumn{4}{c|}{\textbf{Development plan}}\\ \hline
% 		    Spatial Pooling & Spatial Dim & CIFAR10 & CIFAR100 & OOD  \\
% 		  %  \cmidrule(r){2-3} \cmidrule(r){4-5}
% 		    \hlineB{2.5}
%             Dense & $H \times W$ & 95.76 & 74.5 & 94.20\\ 
%             Reduced Dense & $2 \times 2$ & 95.79 & 76.05 & 94.70 \\
%             GAP & $1 \times 1$ & \textbf{95.96} & \textbf{76.36} & \textbf{94.90} \\
%             \hline

%             \hlineB{2.5}
% 		\end{tabular}
% 	}
% % 	\vspace{0.2cm}
% 	\caption{Comparison of three different spatial feature aggregation methods for LoRot-E. Reported results are the  classification accuracies (\%) on CIFAR datasets and AUROC scores (\%) for OOD. Note that, $H$ and $W$ indicate the width and height of the features, respectively.}
% 	\label{table_GAP}
% % 	\vspace{-0.5cm}
% \end{table*}
% \endgroup
\section{Spatial Pooling Methods for LoRot-E}
\label{section:spatialpooling}

We basically use a global average pooling (GAP) layer to spatially aggregate the final convolution layer features for the primary and pretext classifiers in LoRot. However, the task of LoRot-E explicitly includes localizing the rotated region within the image, since it requires predicting the index of the rotated quadrant. Therefore, there are more possible choices to aggregate the final convolution layer features for the pretext task to maintain the spatial information.

We investigate three approaches to figure out the proper spatial pooling method: Dense, Reduced Dense, and GAP.
First, we can keep all the spatial dimensions of the features, called Dense. This way does not sacrifice any spatial information but requires a bunch of additional parameters for the pretext classifier. Second, we can reduce the spatial dimensions to $2 \times 2$, called Reduced Dense. We think $2 \times 2$ is the minimum resolution for LoRot-E since it uses a $2 \times 2$ grid layout as default. For example, when we have $4 \times 4$ feature maps, we can reduce the dimensions into $2 \times 2$ by the average pooling. Third, we can collapse the spatial dimensions to $1 \times 1$, called GAP, used in the paper. As GAP collapses all the spatial dimensions, the model may not be able to localize the quadrant with rotation.
However, we claim that the self-supervision task is still solvable since the features encode the information of the absolute position thanks to the zero paddings~\cite{islam2020much}. Moreover, it is not necessary to introduce additional parameters for the pretext classifier, which avoids the computational overheads.

To validate the effects of each spatial pooling methods, we report the experimental results of three settings for image classification with CIFAR10/100 and OOD in Tab.~\ref{table_GAP}. We use ResNet50 for classification on CIFAR10/100, and ResNet18 for OOD as did in the paper.
% Also, we report the averaged scores for all experiments as Tab.~\ref{table_patchsize} and Tab.~\ref{table_gridsize}
All results are averaged over three or five trials for classification and OOD, respectively.
Note that, the reported AUROC is the averaged score over all out-distribution datasets described in Sec.~\ref{supp:implementationdetails}.
% Also, we report the averaged AUROC score of the datasets for OOD.
Interestingly, GAP consistently outperforms Dense and Reduced Dense. We conjecture that the learned features have information for the localization within their channels.
Moreover, the additional parameters can cause the over-fitting or grant too much weights on the pretext task. As a result, GAP is a proper spatial aggregation method for LoRot-E with higher performances and less computational cost.

% In this work, our ResNet model has global average pooling layer (GAP).
% Then, one might feel this is somewhat weird for PatchRot-G which has the task to predict the location.
% However, as convolutional neural networks are now proven to be capable of encoding spatial information with the padding~\cite{islam2020much}, it is very natural, we think, and we further support this idea with our experiments.
% Specifically, we tested our methods with and without GAP.
% 1$\times$1 GAP in Tab.~\ref{table_GAP} is the original ResNet.
% NO GAP and K$\times$K GAP both maintain the original network that the primary classifier is attached to GAP but add an additional classifier before GAP or after another GAP that keeps the spatial information in K$\times$K shape.
% In Tab.~\ref{table_GAP}, we can see that results not only differ not much from one another but the original ResNet 
% has slightly better performances.
% From this experiment, we believe that convolution layers can track spatial information and this even makes our self-supervision more light-weighted.

% However, as we wrote in the paper, auxiliary self-supervision should be light-weight.
% Thus, in this section, we compare the our results to other possible methods that do not use global average pooling layer to average all spatial information.

% \begingroup
% \setlength{\tabcolsep}{12pt} % Default value: 6pt
% \renewcommand{\arraystretch}{1.2} % Default value: 1

\Skip{
        Basically, we use a global average pooling layer (GAP) to aggregate the features of the final convolution layer for the primary and pretext classifiers in PatchRot.
        However, the task of PatchRot-G involves localizing the rotated region in the image since it requires predicting the index of the quadrant with the rotation.
        Therefore, there are more possible choices to aggregate the features of the final convolution layer for the pretext task to maintain the spatial information.
        
        In this paper, we propose three approaches to figure out the proper aggregation method: Dense, Reduced Dense, and GAP.
        Firstly, we can keep all the spatial dimensions of the features, called Dense.
        This way does not sacrifice any spatial information, but causes a bunch of additional parameters to the pretext classifier.
        Secondly, we can reduce the spatial dimensions to $2 \times 2$, called Reduced Dense.
        We think $2 \times 2$ is the minimum resolution for PatchRot-G since it uses a $2 \times 2$ grid as default.
        For example, when we have $4 \times 4$ feature maps, we can reduce the dimensions into $2 \times 2$ by average pooling.
        Lastly, we can collapse the spatial dimensions to $1 \times 1$, called GAP.
        This way is the same as that of PatchRot and the primary task.
        As GAP collapses all the spatial dimensions, the model may not be able to localize the quadrant with rotation.
        However, we claim that it can solve the pretext task since the features encode the information of the absolute position thanks to the zero paddings~\cite{islam2020much}.
        As so, it is not necessary to introduce additional parameters to the pretext classifier, which results in computational efficiency.
        
        To validate the effects of each aggregation way, we run experiments on three settings for image classification with CIFAR10/100 and OOD.
        We use ResNet50 for classification on CIFAR10/100, and ResNet18 for OOD as done in the paper.
        % Also, we report the averaged scores for all experiments as Tab.~\ref{table_patchsize} and Tab.~\ref{table_gridsize}
        All results are averaged on three or five trials for classification and OOD, respectively.
        Be aware that the reported AUROC is the averaged score over all out-distribution datasets mentioned in Sec.~\ref{supp:implementationdetails}.
        % Also, we report the averaged AUROC score of the datasets for OOD.
        Tab.~\ref{table_GAP} shows the results of three aggregation approaches.
        Interestingly, GAP consistently outperforms Dense and Reduced Dense.
        We conjecture that there is no need to keep the spatial dimensions since the features can handle the information for localizing in their channels.
        Moreover, the additional parameters might cause over-fitting or grant too much weight on the pretext task that can degrade the performance.
        Therefore, GAP is the proper aggregation method with better performance and less computational cost.
        
        % In this work, our ResNet model has global average pooling layer (GAP).
        % Then, one might feel this is somewhat weird for PatchRot-G which has the task to predict the location.
        % However, as convolutional neural networks are now proven to be capable of encoding spatial information with the padding~\cite{islam2020much}, it is very natural, we think, and we further support this idea with our experiments.
        % Specifically, we tested our methods with and without GAP.
        % 1$\times$1 GAP in Tab.~\ref{table_GAP} is the original ResNet.
        % NO GAP and K$\times$K GAP both maintain the original network that the primary classifier is attached to GAP but add an additional classifier before GAP or after another GAP that keeps the spatial information in K$\times$K shape.
        % In Tab.~\ref{table_GAP}, we can see that results not only differ not much from one another but the original ResNet 
        % has slightly better performances.
        % From this experiment, we believe that convolution layers can track spatial information and this even makes our self-supervision more light-weighted.

        % However, as we wrote in the paper, auxiliary self-supervision should be light-weight.
        % Thus, in this section, we compare the our results to other possible methods that do not use global average pooling layer to average all spatial information.
        
        \begingroup
        \setlength{\tabcolsep}{12pt} % Default value: 6pt
        \renewcommand{\arraystretch}{1.2} % Default value: 1
        \begin{table}[h]
        % 	\vspace{0.2cm}
        	\centering
        	{\small
        		\begin{tabular}{l c | c c c}
        		    \hlineB{2.5}
        		  %  \multicolumn{5}{|c|}{\textbf{Review of learning needs}} & \multicolumn{4}{c|}{\textbf{Development plan}}\\ \hline
        		    Pooling Methods & Spatial Dimensions & CIFAR10 & CIFAR100 & OOD  \\
        		  %  \cmidrule(r){2-3} \cmidrule(r){4-5}
        		    \hlineB{2.5}
                    Dense & $H \times W$ & 95.76 & 74.5 & 94.2\\ 
                    Reduced Dense & $2 \times 2$ & 95.79 & 76.05 & 94.70 \\
                    Global Average Pooling & $1 \times 1$ & \textbf{95.96} & \textbf{76.36} & \textbf{94.90} \\
                    \hline

                    \hlineB{2.5}
        		\end{tabular}
        	}
        	\vspace{0.2cm}
        	\caption{Comparison over different types of feature aggregation methods. Experiments are conducted to measure classification accuracy (\%) on CIFAR and AUROC (\%) for OOD. $H$, and $W$ indicate the width and height of the features, respectively.}
        	\label{table_GAP}
        % 	\vspace{-0.5cm}
        \end{table}
        \endgroup
        
}

%% file: Supp_SupCLRFullOOD.tex
\begingroup
\setlength{\tabcolsep}{3.2pt} % Default value: 6pt
\renewcommand{\arraystretch}{1.1} % Default value: 1
\begin{table*}[t]
% 	\vspace{0.2cm}
	\centering
		\caption{Full AUROC (\%) results of the averaged OOD scores reported in Tab.~7 in the paper. Models are trained with CIFAR10 dataset and evaluated on each out distribution dataset listed in the table. IN denotes ImageNet.}
	\label{table_full_ood_supcon}
	{\scriptsize
		\begin{tabular}{l| c | c c c c c c | c}
		    \hlineB{2.5}
		  %  \multicolumn{2}{c}{} & \multicolumn{6}{c}{In Distribution : CIFAR10} \\
            % \cmidrule(r){3-8}
            % \multicolumn{2}{c}{} & \multicolumn{6}{c}{Out Distribution} \\ 
            Method & Model & SVHN & LSUN & IN & LSUN (FIX) & IN (FIX) & CIFAR-100 & Avg \\
			\hlineB{2.5}
			SupCLR~\cite{khosla2020supervised} & ResNet50 & \underline{98.6} & 97.1 & 96.2 & 97.3 & 97.1 & 95.6 & 96.98 \\
			\hline
 			+ Rot (MT)~\cite{gidaris2018unsupervised} & ResNet50 & 98.2 & 98.0 & 97.4 & 95.7 & 95.0 & 93.4 & 96.28 \\
 			+ Rot (PT)~\cite{gidaris2018unsupervised}& ResNet50 & 98.0 & 98.2 & \underline{97.9} & 96.1 & 96.3 & 94.9 & 96.90 \\
 			\hline
 			+ LoRot-I & ResNet50 & \textbf{99.1} & \underline{98.6} & \underline{97.9} & \textbf{98.0} & \textbf{97.7} & \textbf{96.4} & \textbf{97.95} \\
 			+ LoRot-E & ResNet50 & \textbf{99.1} & \textbf{98.9} & \textbf{98.4} & \underline{97.8} & \underline{97.3} & \underline{96.0} & \underline{97.92} \\
            \hlineB{2.5}
		\end{tabular}
	}
	\vspace{-0.3cm}

% 	\vspace{-0.7cm}
\end{table*}
% % \begingroup
% % \setlength{\tabcolsep}{2.2pt} % Default value: 6pt
% % \renewcommand{\arraystretch}{1.2} % Default value: 1
% \begin{table*}[t]
% % 	\vspace{0.2cm}
% 	\centering
% 	{\small
% 		\begin{tabular}{l| c | c c c c c c | c}
% 		    \hlineB{2.5}
% 		    \multicolumn{2}{c}{} & \multicolumn{6}{c}{In Distribution : CIFAR10} \\
%             \cmidrule(r){3-8}
%             \multicolumn{2}{c}{} & \multicolumn{6}{c}{Out Distribution} \\ 
%             Method & Model & SVHN & LSUN & ImageNet & LSUN (FIX) & ImageNet (FIX) & CIFAR-100 & Avg \\
% 			\hlineB{2.5}
% 			SupCLR~\cite{khosla2020supervised} & ResNet50 & \underline{98.6} & 97.1 & 96.2 & 97.3 & 97.1 & 95.6 & 96.98 \\
% 			\hline
%  			+ Rot (MT)~\cite{gidaris2018unsupervised} & ResNet50 & 98.2 & 98.0 & 97.4 & 95.7 & 95.0 & 93.4 & 96.28 \\
%  			+ Rot (PT)~\cite{gidaris2018unsupervised}& ResNet50 & 98.0 & 98.2 & \underline{97.9} & 96.1 & 96.3 & 94.9 & 96.90 \\
%  			\hline
%  			+ LoRot-I & ResNet50 & \textbf{99.1} & \underline{98.6} & \underline{97.9} & \textbf{98.0} & \textbf{97.7} & \textbf{96.4} & \textbf{97.95} \\
%  			+ LoRot-E & ResNet50 & \textbf{99.1} & \textbf{98.9} & \textbf{98.4} & \underline{97.8} & \underline{97.3} & \underline{96.0} & \underline{97.92} \\
%             \hlineB{2.5}
% 		\end{tabular}
% 	}
% % 	\vspace{0.15cm}
% 	\caption{Raw results of the averaged OOD scores reported in Table.~7 of the paper.}
% 	\label{table_full_ood_supcon}
% % 	\vspace{-0.7cm}
% \end{table*}
% % \endgroup
\begingroup
\setlength{\tabcolsep}{7.2pt} % Default value: 6pt
\renewcommand{\arraystretch}{1.} % Default value: 1
\begin{table*}[t]
% 	\vspace{0.2cm}
    \centering
        \caption{Imbalanced classification accuracy (\%) on CIFAR10/100. Experiments are conducted with the supervised baseline. The table demonstrates the complementary benefits of LoRot in the data imbalance settings. }
    \label{table_IM_ce}
    {\small
        \begin{tabular}{l | c c c | c c c}
            \hlineB{2.5}
          %  \multicolumn{5}{|c|}{\textbf{Review of learning needs}} & \multicolumn{4}{c|}{\textbf{Development plan}}\\ \hline
            \multicolumn{1}{c}{} & \multicolumn{3}{c}{Imbalanced CIFAR10} & \multicolumn{3}{c}{Imbalanced CIFAR100} \\
            % \cmidrule(r){1-4} \cmidrule(r){4-7}
            \hline
            Imbalance Ratio & 0.01 & 0.02 & 0.05 & 0.01 & 0.02 & 0.05  \\
            \hlineB{2.5}
        
            Baseline & 70.36 & 78.06 & 83.42 & 38.32 & 43.80 & 51.00\\
            + Rot (DA) & 64.78 & 70.19 & 77.41 & 35.15 & 38.53 & 50.99 \\
            + Rot (MT) & 66.01 & 71.75 & 78.18 & 35.76 & 39.08 & 46.24 \\
            + Rot (PT) & 71.75 & 76.31 & 83.68 & 38.91 & 43.62 & 50.99 \\
            + LoRot-I & \underline{74.79} & \underline{80.40} & \underline{85.42} & \underline{39.42} & \underline{45.71} & \underline{53.16}\\
            + LoRot-E & \textbf{77.32} & \textbf{80.67} & \textbf{85.67} &  \textbf{41.99} & \textbf{47.72} & \textbf{54.97} \\
            \hlineB{2.5}
        \end{tabular}
    }
    \vspace{-0.3cm}

\end{table*}
\endgroup
\section{OOD Detection with Contrastive Learning}
\label{OODwithCSI}
In Tab.~\ref{table_full_ood_supcon}, we report raw individual results of the averaged OOD detection scores shown in \textcolor{blue}{Tab.~7} in the main paper. We observe performance gains for all datasets when LoRot used in conjunction with supervised contrastive learning~\cite{khosla2020supervised}. 
On the other hand, the original rotation task hardly improves and even degrades the performance of SupCLR in either multi-tasking (MT) or parallel-task learning (PT) strategy.
Particularly, although the original rotation task enhances the performance in LSUN- and ImageNet-resize datasets, it shows the slight degradations for other OOD datasets.

\Skip{
        We report raw values of the averaged OOD results in Table.~7 in Sec.~4 of the paper about the contrastive learning method SupCLR~\cite{khosla2020supervised} combined with ours.
        As the baseline with ResNet50, itself already has a strong performance, the results do not show much difference.
        Still, in Table.~\ref{table_full_ood_supcon}, PatchRot improves the ability of SupCLR to detect OOD samples, achieving almost 98\% on average.
        % In contrast, when other self-supervisions such as rotation is applied to SupCLR, it deteriorates the performance in both forms: multi-tasking learning and parallel-task learning.
        In contrast, when rotation is applied to SupCLR in either form of multi-tasking (MT) or parallel-task learning (PL), it deteriorates the performance in SupCLR.

        \label{subsection:OODExperiments}
        \begingroup
        \setlength{\tabcolsep}{2.2pt} % Default value: 6pt
        \renewcommand{\arraystretch}{1.2} % Default value: 1
        \begin{table}[h]
        % 	\vspace{0.2cm}
        	\centering
        	{\small
        		\begin{tabular}{l| c | c c c c c c | c}
        		    \hlineB{2.5}
        		    \multicolumn{2}{c}{} & \multicolumn{6}{c}{In Distribution : CIFAR10} \\
                    \cmidrule(r){3-8}
                    \multicolumn{2}{c}{} & \multicolumn{6}{c}{Out Distribution} \\ 
                    Method & Model & SVHN & LSUN & ImageNet & LSUN (FIX) & ImageNet (FIX) & CIFAR-100 & Avg \\
        			\hlineB{2.5}
        			SupCLR~\cite{khosla2020supervised} & ResNet50 & \underline{98.6} & 97.1 & 96.2 & 97.3 & 97.1 & 95.6 & 96.98 \\
        			\hline
         			+ Rot (MT)~\cite{gidaris2018unsupervised} & ResNet50 & 98.2 & 98.0 & 97.4 & 95.7 & 95.0 & 93.4 & 96.28 \\
         			+ Rot (PT)~\cite{gidaris2018unsupervised}& ResNet50 & 98.0 & 98.2 & \underline{97.9} & 96.1 & 96.3 & 94.9 & 96.90 \\
         			\hline
         			+ PatchRot (MT)& ResNet50 & \textbf{99.1} & \underline{98.6} & \underline{97.9} & \textbf{98.0} & \textbf{97.7} & \textbf{96.4} & \textbf{97.95} \\
         			+ PatchRot-G (MT)& ResNet50 & \textbf{99.1} & \textbf{98.9} & \textbf{98.4} & \underline{97.8} & \underline{97.3} & \underline{96.0} & \underline{97.92} \\
                    \hlineB{2.5}
        		\end{tabular}
        	}
        	\vspace{0.15cm}
        	\caption{Full AUROC (\%) scores for distinguishing in- and out-of-distribution data for image classification model with ResNet50. Classification accuracy is in the paper. Check Table. 7 in Sec. 4.}
        	\label{table_full_ood_supcon}
        	\vspace{-0.7cm}
        \end{table}
        \endgroup
        
}

%% file: Supp_ImbalancedClassification.tex
\section{Imbalanced classification with the baseline}
% \paragraph{Imbalanced image classification}
\label{imbalanced_baseline}
In \textcolor{blue}{Tab.~4} of the main paper, we explored complementary benefits of LoRot in the imbalanced image classification with LDAM-DRW\cite{cao2019learning} under varying imbalanced scenarios.
% Hereby, we additionally show that LoRot is not only 
% In the main paper Tab. 4, we reported the experiments with LDAM-DRW\cite{cao2019learning} to show the complementary benefits on imbalanced classification.
In this supplementary report, we additionally show LoRot's compatibility to the baseline model.
%state of the art method but also with the baseline. 
As shown in Tab.~\ref{table_IM_ce}, we observe that LoRot provides complementary benefits with the baseline model. %effective for dealing with class-imbalance problems.
Note that, LoRot-E combined with the baseline even outperforms LDAM-DRW in four out of six scenarios.

%% file: Supp_OODanalysis.tex
\begin{figure*}[t!]
    \centering
    \includegraphics[scale=0.65]{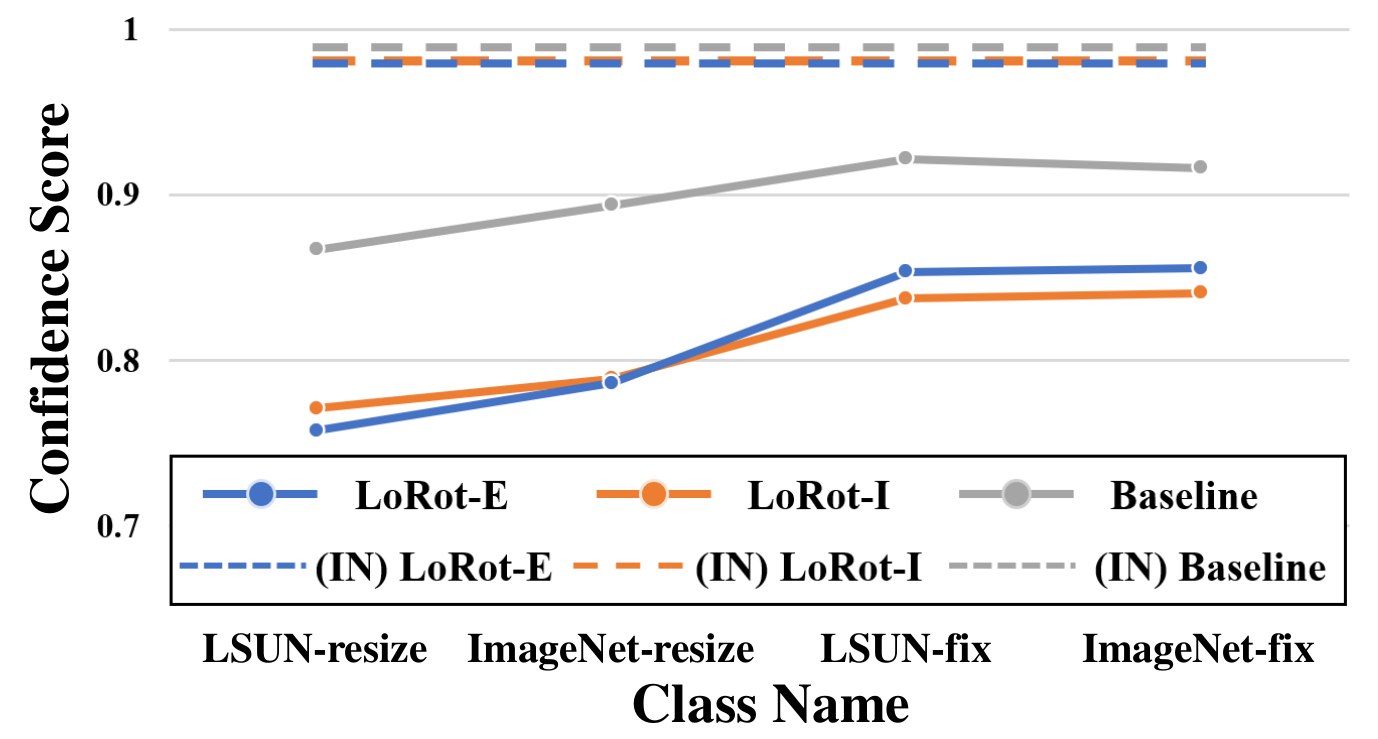}
    \vspace{-0.2cm}
    \caption{Dataset-wise averaged confidence scores for in- and out-of-distribution data of the baseline and LoRot.
    As used in \textcolor{blue}{Fig.~5} in the paper, dotted lines are the averaged confidence scores of in-distribution (IN-) dataset (CIFAR10) and solid lines represent the confidence scores (y-axis) for each dataset (x-axis). These results demonstrate that the benefits of LoRot are consistent across datasets.}
    %the model's unknown detecting capability across novel classes.}
    %Class-by-class performance analysis with confidence scores from the baseline and LoRot.
    \label{figure_OOD_conf}
    
\end{figure*}
\begin{figure*}[t]
    \centering
    \includegraphics[scale=0.4]{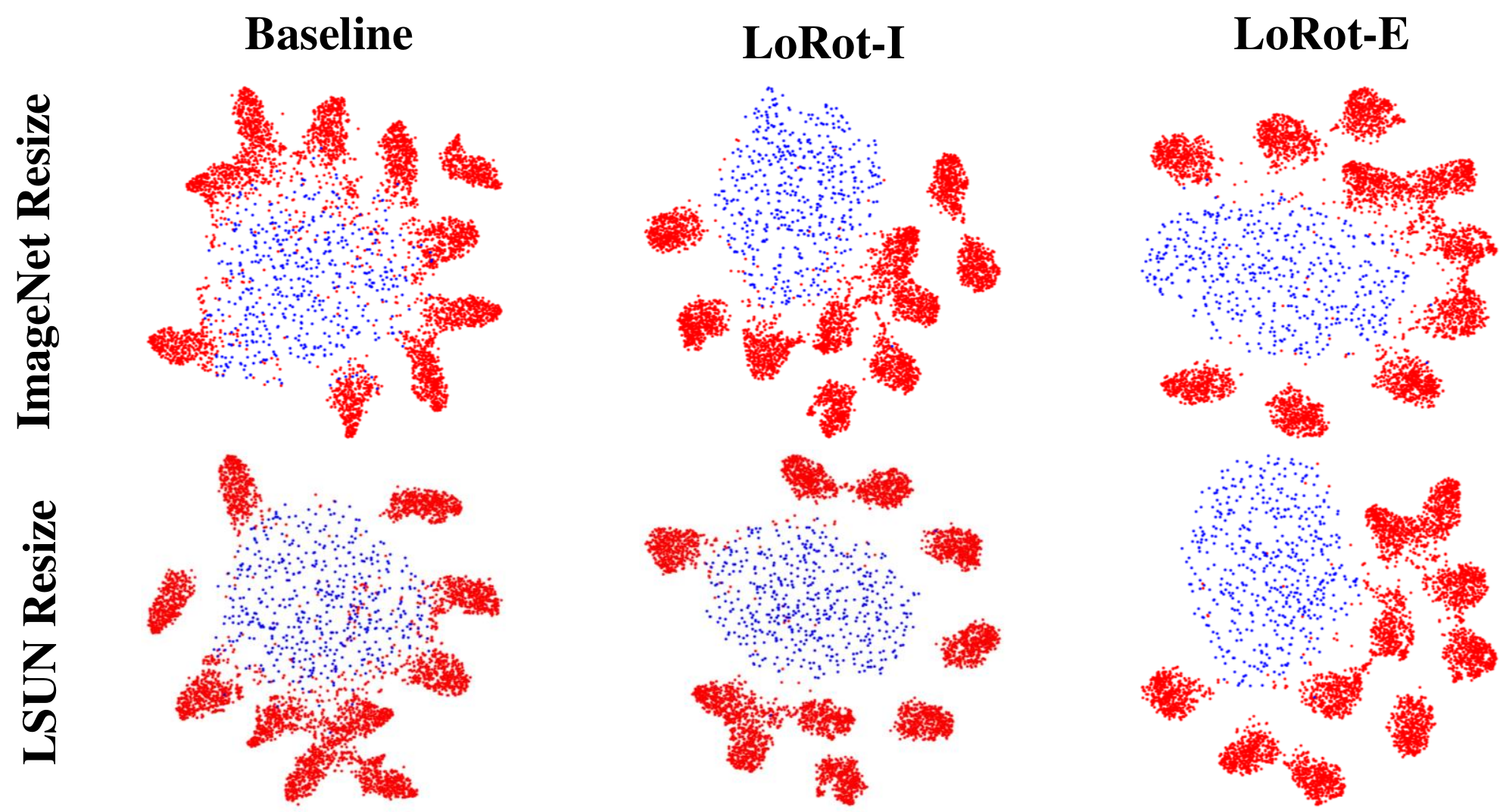}

    \caption{t-SNE visualization for the baseline, LoRot-I, and LoRot-E on OOD detection benchmark. Dataset for blue dots (OOD dataset) is indicated on the left. }
    \label{figure_OOD_TSNE}
    \vspace{-0.2cm}
\end{figure*}
\section{Further study of LoRot on other datasets in OOD detection}
\label{section:OODanalysis}
We further show the results on other datasets observing why LoRot is effective in detecting unknown samples.
To be specific, we measured the average confidence scores for LSUN and ImageNet datasets in Fig.~\ref{figure_OOD_conf} since their class labels are not available.
These results show that LoRot consistently improves the robustness of the classifier by encouraging the model to yield lower confidence scores for unknown OOD datasets.
%datasets where the labels are provided, while the average confidence scores over whole dataset are measured for LSUN and ImageNet
%, and average confidence scores 
%that provide the class labels (SVHN and CIFAR100) and otherwise, averaged confidence scores per dataset (LSUN and ImageNet). 
% Therefore, we claim that a clear tendency observed in the paper that both the LoRot-I and LoRot-E effectively lower the confidence scores for OOD datasets is consistent across datasets.
Furthermore, plotted t-SNE in Fig.~\ref{figure_OOD_TSNE} also implies how better separation between in- and out-distribution datasets are achieved on ImageNet and LSUN dataset. 
Note that red clusters represent each class in in-distribution dataset (CIFAR10).

% \begin{table}[t]
%     \centering
%     {\small
% 		\begin{tabular}{l | c c c}
% 		    \hlineB{2.5}
% 		    \multicolumn{1}{c}{} & FS & Training Time & Inference Time \\
% 		    \hlineB{2.5}
% 		    LoRot-I & 4309 & 56 & \\
% 		    LoRot-E & 4499 & 56 & \\
%             Rotation~\cite{hendrycks2019using} & - & 111  \\
%             CSI~\cite{hendrycks2019using} & 125275(+2635) & \\
            
%             \hlineB{2.5}
% 		\end{tabular}
% 	}
% 	\caption{}
% 	\label{table_OOD_time}
% \end{table}

%% file: Supp_implementationDetails.tex
\section{Implementation Details}
\label{supp:implementationdetails}
% We provide implementation details for the conducted experiments in the paper to ease the reproduction of the results.

% \textbf{Open-Set Recognition.}
% We train the model for 200 epochs with a batch size of 64.
% Our encoder is consisted of 9 convolution layers with additional classifier to perform multi-task learning.
% Adam~\cite{kingma2015adam} optimizer with the initial learning rate of 0.001 is used and the learning rate is dropped by the factor of 0.1 at 100 epoch. For the MNIST dataset, since it converges at 99\% for both accuracy and AUROC in earlier epochs, we only train 50 epochs.

\textbf{OOD Detection.}
For the OOD detection, we use the ResNet18 architecture as the backbone for a fair comparison against the reported performances in the literature. Therefore, some of the results are reproduced based on their original implementations to unify the backbone network.
We deploy the Adam~\cite{kingma2015adam} optimizer with a batch size 64, and a learning rate of 0.001.
We train the network for 100 epochs and the learning rate is decayed at the middle point of learning by the factor of 0.1.
For Rotations~\cite{hendrycks2019using} and SLA+SD~\cite{lee2020self}, we set the batch size to 128 since it shows the better performances as used in their original papers. As described in the paper Sec. 4.1.1, we use CIFAR-10 as in-distribution data, while SVHN~\cite{netzer2011reading}, the resized versions of ImageNet and LSUN~\cite{liang2017enhancing}, the fixed versions of ImageNet and LSUN~\cite{tack2020csi}, and CIFAR-100~\cite{krizhevsky2009learning} are treated as the out-of-distribution data.
% To measure the performance of Patch-Rot, we utilize the KL-divergence between the softmax predictions and the uniform distribution as in \cite{hendrycks2018deep, hendrycks2019using}.
% However, we use the softmax predictions for SLA+SD~\cite{lee2020self} and CutMix~\cite{yun2019cutmix} as the softmax results fit better with their methods. The performances for SupCLR~\cite{khosla2020supervised} and CSI~\cite{tack2020csi} are from their papers. 
% % In OOD detection setting, we set our hyper-parameter $\gamma$ to 0.1 for all experiments since it shows the best performance (We also test our model using 0.1 and 0.5 for $\gamma$, but there is no significant performance gain). For the fairness, we averaged 5 runs for all datasets for the final results.

\textbf{Imbalanced Classification.}
For a fair comparison, we use the ResNet-32 architecture as the backbone network and follow the settings of the baseline\cite{cao2019learning}. 
We set the batch size to 128, and the initial learning rate to 0.1 which is dropped by 0.01 at the 160-th, and 180-th epochs.
SGD is used for the optimizer with a momentum of 0.9, weight decay of 2$\times$ 10$^{-4}$.

\textbf{Adversarial Perturbations}
Following previous work\cite{hendrycks2019using}, we adopted wide ResNet 40-2\cite{zagoruyko2016wide} architecture as the backbone network.
For more details, we utilize SGD optimizer with Nesterov momentum of 0.9 and a batch size of 128. 
Also, we use an initial learning rate of 0.1 with cosine learning rate schedule\cite{loshchilov2016sgdr} and weight decay of 5$\times10^4$.

\textbf{Standard Image Classification.}
% For the classification experiments, we train the model for 500 epochs with a batch size of 512 and learning rate of 0.8.
% During training, the learning rate is decayed by the cosine decay scheduler.
% For SLA~\cite{lee2020self}, we followed all the settings and parameters described in the paper, except that we used cosine decay scheduler and set the number of training epochs to 125 since they forward 4 times more samples in each epoch for CIFAR dataset. For all methods, we use stochastic gradient descent (SGD) with the momentum of 0.9 and weight decay of $1\times 10^{-4}$.
For ImageNet classification, we train the model for 300 epochs with a batch size of 256 and initial learning rate of 0.1.
During the training, the learning rate is decayed at every 75 epochs with the decaying factor of 0.1.
The same implementation details are also applied for Tab. 7, where we describe the complementary benefits to data augmentation methods, e.g., Mixup\cite{zhang2017mixup}, AutoAugment\cite{cubuk2019autoaugment}, and RandAugment\cite{cubuk2020randaugment}.
\newline
For experiments regarding contrastive learning, we follow the protocols from the SupCLR\cite{khosla2020supervised} except for the batch size due to lack of GPU memory. 
Specifically, we train ResNet50 for 1000 epochs with the batch size of 512. The initial learning rate is set to 0.05 and is decayed by cosine decay scheduler.
Then, with the learning rate of 5, the classifier is finetuned for evaluation.
% we used 256 for the batch size and 0.1 for the learning rate and trained the model for 300 epochs. We decayed the learning rate by the factor of 0.1 in every 75 epochs.

\textbf{Transfer Learning}
For instance segmentation, we use real-time SOLOv2\cite{wang2020solov2} model where the number of convolution layers in the prediction head is reduced to two and the input shorter side is 448. We train the model with the 3x schedule as reported in their paper. 
To train RetinaNet~\cite{lin2017focal}, we use the 1x schedule. 
% For details, check their official repository \url{github.com/facebookresearch/detectron2}. 
For both experiments, we test on MS COCO 2017 dataset~\cite{lin2014microsoft} with the ResNet50 architecture as the backbone network.